\documentclass[journal]{IEEEtran}

\usepackage{makecell,threeparttable,float,algorithmic,palatino,epsfig,latexsym,cite,graphicx,amsthm,amsmath,amssymb,amsfonts,multirow,booktabs,color,soul}
\usepackage[linesnumbered,ruled,vlined]{algorithm2e}
\usepackage[caption=false,font=footnotesize]{subfig}
\definecolor{hl}{rgb}{0.75,0.75,0.75}
\sethlcolor{hl}

\def\MR2{\multirow{2}{*}}
\theoremstyle{plain}
\newtheorem{theorem}{Theorem}
\newtheorem{definition}{Definition}
\newtheorem{corollary}{Corollary}

\begin{document}

\title{Principled Design of Translation, Scale, and Rotation Invariant Variation Operators\\for Metaheuristics}
\author{Ye Tian,
        Xingyi Zhang,
        Cheng He,
        Kay Chen Tan,
        and
        Yaochu Jin
\thanks{Y. Tian is with the Key Laboratory of Intelligent Computing and Signal Processing of Ministry of Education, Institutes of Physical Science and Information Technology, Anhui University, Hefei 230601, China (email: field910921@gmail.com).}
\thanks{X. Zhang is with the Key Laboratory of Intelligent Computing and Signal Processing of Ministry of Education, School of Computer Science and Technology, Anhui University, Hefei 230601, China (email: xyzhanghust@gmail.com).}
\thanks{C. He is with the Guangdong Provincial Key Laboratory of Brain-inspired Intelligent Computation, Department of Computer Science and Engineering, Southern University of Science and Technology, Shenzhen 518055, China (email: chenghehust@gmail.com).}
\thanks{K. C. Tan is with the Department of Computing, The Hong Kong Polytechnic University, Hong Kong SAR (email: kctan@polyu.edu.hk).}
\thanks{Y. Jin is with the Department of Computer Science, University of Surrey, Guildford, Surrey, GU2 7XH, U.K. (email: yaochu.jin@surrey.ac.uk).}
}



\maketitle

\begin{abstract}
In the past three decades, a large number of metaheuristics have been proposed and shown high performance in solving complex optimization problems. While most variation operators in existing metaheuristics are empirically designed, this paper aims to design new operators automatically, which are expected to be search space independent and thus exhibit robust performance on different problems. For this purpose, this work first investigates the influence of translation invariance, scale invariance, and rotation invariance on the search behavior and performance of some representative operators. Then, we deduce the generic form of translation, scale, and rotation invariant operators. Afterwards, a principled approach is proposed for the automated design of operators, which searches for high-performance operators based on the deduced generic form. The experimental results demonstrate that the operators generated by the proposed approach outperform state-of-the-art ones on a variety of problems with complex landscapes and up to 1000 decision variables.

\end{abstract}

\section{Introduction}

\IEEEPARstart{M}{etaheuristics} have shown effectiveness in solving the complex optimization problems from various fields, such as manufacturing \cite{Alam2003Process}, scheduling \cite{Potvin1996Vehicle}, bioinformatics \cite{App-ModuleIdentification4}, and economics \cite{App-PortfolioOptimization4}. In contrast to mathematical programming methods that require an explicit objective function \cite{Operator-GD}, metaheuristics provide a high-level methodology solving problems in a black-box manner. So far, there have been a variety of algorithms inspired by the biological mechanisms in natural evolution, such as genetic algorithms (GA) \cite{Holland1992Adaptation}, differential evolution (DE) \cite{Operator-DE}, evolution strategies \cite{algorithm-CMAES}, and evolutionary programming \cite{Operator-FEP}. Moreover, many swarm intelligence based algorithms have also been proposed, including particle swarm optimization (PSO) \cite{Operator-PSO}, ant colony optimization \cite{algorithm-ACO}, artificial bee colony algorithms \cite{Operator-ABC}, among many others.

\IEEEpubidadjcol

Existing metaheuristics exhibit quite different search behaviors and optimization performance, which are mainly determined by the manually designed variation operators \cite{Survey2,Okabe2005Theoretical,Survey1}, i.e., the strategies for receiving the decision vectors of parents and generating the decision vectors of offspring solutions. For example, the GA was designed according to the evolution theory and law of inheritance, which generates offspring by the crossover between two parents and the mutation on a single offspring solution. The crossover and mutation operators provide a powerful exploration ability \cite{Eiben1998Evolutionary}, making GA good at handling multimodal landscapes \cite{Su2020Non}. The DE mutates each solution according to the weighted difference between the other two solutions, which holds a good performance on problems with complicated variable linkages \cite{MOEA/D-DE}. The covariance matrix adaptation evolution strategy (CMA-ES) generates new solutions by sampling a multivariate normal distribution model adaptively learned from the population, showing high performance on many real-world applications \cite{Rodemann2018Industrial}. Inspired by the choreography of bird flocking, the PSO updates each particle according to its personal best particle and the global best particle, which has a high speed of convergence \cite{MOPSO}.

Among the superiorities in existing variation operators, the independence of search space is crucial to the robustness and generalization of metaheuristics, since metaheuristics do not rely on specific characteristics of problems. A search space independent operator holds the same performance on a problem with arbitrary search space transformations, including translation (i.e., $x^\prime=x+b$), scaling (i.e., $x^\prime=ax$), and rotation (i.e., $\mathbf{x}^\prime=\mathbf{x}M$). Existing studies have shed some light on the sufficient conditions of achieving these invariance properties. For instance, CMA-ES is scale invariant since its step-sizes are set proportionally to the distance to the optimum found so far \cite{Jebalia2011Log}, and the mutation operator of DE is rotation invariant since it is the weighted sum of multiple parents \cite{Caraffini2019Study}. Nevertheless, the mathematical definitions of all the three properties have not been given, and the sufficient and necessary condition of achieving them is still unknown. Therefore, it is difficult to analyze whether an operator is translation, scale, and rotation invariant theoretically, or to consider these properties in designing new operators explicitly.

To address this issue, this work aims to deduce the generic form (i.e., sufficient and necessary condition) of translation, scale, and rotation invariant operators, which can be used to judge the possession of invariance properties and guide the design of new operators. Based on the deduced generic form, this work proposes a principled approach for designing new operators with invariance properties. In contrast to the parameter tuning of metaheuristics \cite{Karafotias2015Parameter,Huang2019Survey}, the off-line recommendation of metaheuristics \cite{Tian2020Recommender}, and the on-line combination of metaheuristics \cite{framework-AutoMOEA}, the proposed approach is not based on any existing metaheuristic but searches for totally new operators. Moreover, the proposed approach does not utilize any existing optimizer or classifier (e.g., F-Race for parameter tuning \cite{Birattari2002Racing}, artificial neural network for metaheuristic recommendation \cite{Rosenblatt1958Perceptron}, and sum-of-ranks multiarmed bandit algorithm for metaheuristic combination \cite{Fialho2002Toward}). By contrast, it is a self-contained approach that can search for new operators by itself. The main components of this work include the following three aspects:
\begin{itemize}
\item\textbf{Theoretical analysis:} To illustrate the importance of translation invariance, scale invariance, and rotation invariance, their effects on the search behavior and performance of metaheuristics are investigated. Then, the sufficient and necessary condition of achieving these properties is mathematically derived, which reveals the generic form of search space independent operators.
\item\textbf{New approach:} A principled approach to automated design of variation operators is proposed, termed AutoV. Based on the deduced generic form of variation operators, AutoV converts the search of high-performance operators into an optimization problem, in which the decision variables are the parameters in the operators. This way, AutoV can solve optimization problems without relying on any existing operators.
\item\textbf{Experimental study:} The variation operator found by AutoV is embedded in a simple evolutionary framework and compared to eight classical or state-of-the-art metaheuristics. The experimental results show that the operator found by AutoV can obtain the best results on various challenging benchmark problems; in particular, it outperforms the winner of the CEC competition that contains multiple operators with complex adaptation strategies. The results indicate that AutoV has the potential to replace the laborious process of manual design of new metaheuristics.
\end{itemize}
The rest of this paper is organized as follows. Section~II analyzes the effects of the three invariance properties, and Section~III deduces their sufficient and necessary condition. Section~IV presents the proposed principled approach, and Section~V gives the experimental studies. Finally, Section~VI concludes this paper.

\section{Effects of Invariance Properties}

This work focuses on the variation operators for the following continuous optimization problem:
\begin{equation}
\begin{aligned}
\min&\ \ \ f(\mathbf{x})\\
{\rm s.\,t.}&\ \ \ \mathbf{l}\leq \mathbf{x}\leq \mathbf{u}
\end{aligned}
\ ,
\end{equation}
where $\mathbf{x}=(x_1,x_2,\dots,x_D)$ is a decision vector denoting a solution for the problem, $\mathbf{l}=(l_1,l_2,\dots,l_D)$ denotes the lower bound, $\mathbf{u}=(u_1,u_2,\dots,u_D)$ denotes the upper bound, and $D$ is the number of decision variables. To solve such problems in a black-box manner, a variety of variation operators have been proposed to generate solutions without using any specific information except for the lower and upper bounds. This section first introduces some representative variation operators, then presents the definitions of translation invariance, scale invariance, and rotation invariance and analyzes their effects by examples and experimental studies.

\subsection{Variation Operators in Metaheuristics}

An operator generally receives the decision vectors of one or more parents, then outputs the decision vectors of one or more offspring solutions. For example, the simulated binary crossover (SBX) operator \cite{Operator-SBX} used in GA uses two parents $\mathbf{x}_1$ and $\mathbf{x}_2$ to generate two offspring solutions $\mathbf{o}_1$ and $\mathbf{o}_2$ each time:
\begin{equation}
\left \{
\begin{aligned}
o_{1d} = 0.5\left[ (1+\beta)x_{1d}+(1-\beta)x_{2d}\right]  \\
o_{2d} = 0.5\left[ (1-\beta)x_{1d}+(1+\beta)x_{2d}\right]  \\
\end{aligned}
\ ,\ 1\leq d \leq D
\right.,
\label{equ:SBX}
\end{equation}
where $o_{1d}$ denotes the $d$-th variable of solution $\mathbf{o}_1$ and $\beta$ is a random number obeying a special distribution. The mutation operator of DE/rand/1/bin \cite{Operator-DE} uses three parents $\mathbf{x}_1$, $\mathbf{x}_2$, and $\mathbf{x}_3$ to generate an offspring solution $\mathbf{o}$ each time:
\begin{equation}
o_{d} = x_{1d} + F\cdot(x_{2d}-x_{3d})\ ,\ 1\leq d \leq D\ ,
\label{equ:DE}
\end{equation}
where $F$ is a parameter controlling the amplification of the difference between $\mathbf{x}_2$ and $\mathbf{x}_3$. In contrast to the random parameter $\beta$ in SBX that varies on each dimension, the parameter $F$ in DE is a predefined constant. The operator of CMA-ES \cite{algorithm-CMAES} generates offspring by sampling a multi-variate normal distribution:
\begin{equation}
\mathbf{o} = \mathbf{x}_m + \sigma\cdot\mathcal{N}(\mathbf{0},C)\ ,
\label{equ:CMAES}
\end{equation}
where $\mathbf{x}_m$ is the weighted sum of all solutions, $\sigma$ is a vector of iteratively updated step-sizes, and $C$ is a covariance matrix updated according to the current population. In general, the initial $\sigma$ can be set to $0.6(\mathbf{u}-\mathbf{l})$ \cite{Igel2007Covariance}. Similar to CMA-ES, the operator of fast evolutionary programming (FEP) \cite{Operator-FEP} generates offspring by sampling a single-variate normal distribution:
\begin{equation}
o_d = x_d + \eta_d\cdot\mathcal{N}(0,1)\ ,\ 1\leq d \leq D\ ,
\label{equ:FEP}
\end{equation}
where $\mathbf{\eta}$ is a vector of self-adaptive standard deviations related to each solution $\mathbf{x}$, whose elements can be initialized to 3 \cite{Operator-FEP}.

It can be found that these operators generate offspring by using distinct formulas. Generally, an operator can be regarded as a function $h(x_{1d},x_{2d},\dots)$ performed on each dimension $d$, where $x_{1d},x_{2d},\dots$ contains the decision variables of parents, the lower bound, and the upper bound. Note that other parameters (e.g., $\beta$ in SBX and $C$ in CMA-ES) are ignored for simplicity. In the following, we investigate how the invariance properties influence the search behavior and performance of operators.

\subsection{Effects of Translation Invariance}

According to \cite{Hansen2011Impacts}, a variation operator $h(x_{1d},x_{2d},\dots)$ is invariant to search space transformation $\mathcal{T}$ means that $h(\mathcal{T}(x_{1d}),\mathcal{T}(x_{2d}),\dots)=\mathcal{T}(h(x_{1d},x_{2d},\dots))$. A translation of the search space can be regarded as an addition of each decision variable, i.e., $\mathcal{T}(x)=x+b$, hence the translation invariance property can be defined as follows:
\begin{definition}[\textbf{Translation invariance}]
A variation operator $h(x_{1d},x_{2d},\dots)$ is translation invariant if and only if
\begin{equation}
h(x_{1d}+b,x_{2d}+b,\dots)=h(x_{1d},x_{2d},\dots)+b
\end{equation}
holds for any real constant $b$.
\end{definition}

It is not difficult to find that all the operators of SBX, DE, CMA-ES, and FEP are translation invariant. In particular, the SBX operator described in (\ref{equ:SBX}) can be rewritten as
\begin{equation}
h_{sbx}(x_{1d},x_{2d})=x_{1d}+0.5(1\pm\beta)(x_{2d}-x_{1d})\ ,
\label{equ:SBX2}
\end{equation}
hence
\begin{equation}
\begin{aligned}
&h_{sbx}(x_{1d}+b,x_{2d}+b)\\
&\qquad=x_{1d}+b+0.5(1\pm\beta)(x_{2d}+b-x_{1d}-b)\\
&\qquad=h_{sbx}(x_{1d},x_{2d})+b
\end{aligned}
\end{equation}
and the operator is translation invariant. By contrast, if the SBX operator is modified to
\begin{equation}
h_{sbx^\prime}(x_{1d},x_{2d})=0.1x_{1d}+0.5(1\pm\beta)(x_{2d}-x_{1d})\ ,
\end{equation}
then
\begin{equation}
\begin{aligned}
&h_{sbx^\prime}(x_{1d}+b,x_{2d}+b)\\
&\qquad=0.1x_{1d}+0.1b+0.5(1\pm\beta)(x_{2d}+b-x_{1d}-b)\\
&\qquad\neq h_{sbx^\prime}(x_{1d},x_{2d})+b
\end{aligned}
\end{equation}
and the operator becomes not translation invariant. Obviously, the modified SBX$^\prime$ operator is likely to evolve the population toward the origin.

\begin{figure}[!t]
  \centering
  \subfloat[SBX based GA (translation invariant)]{\includegraphics[width=0.5\linewidth]{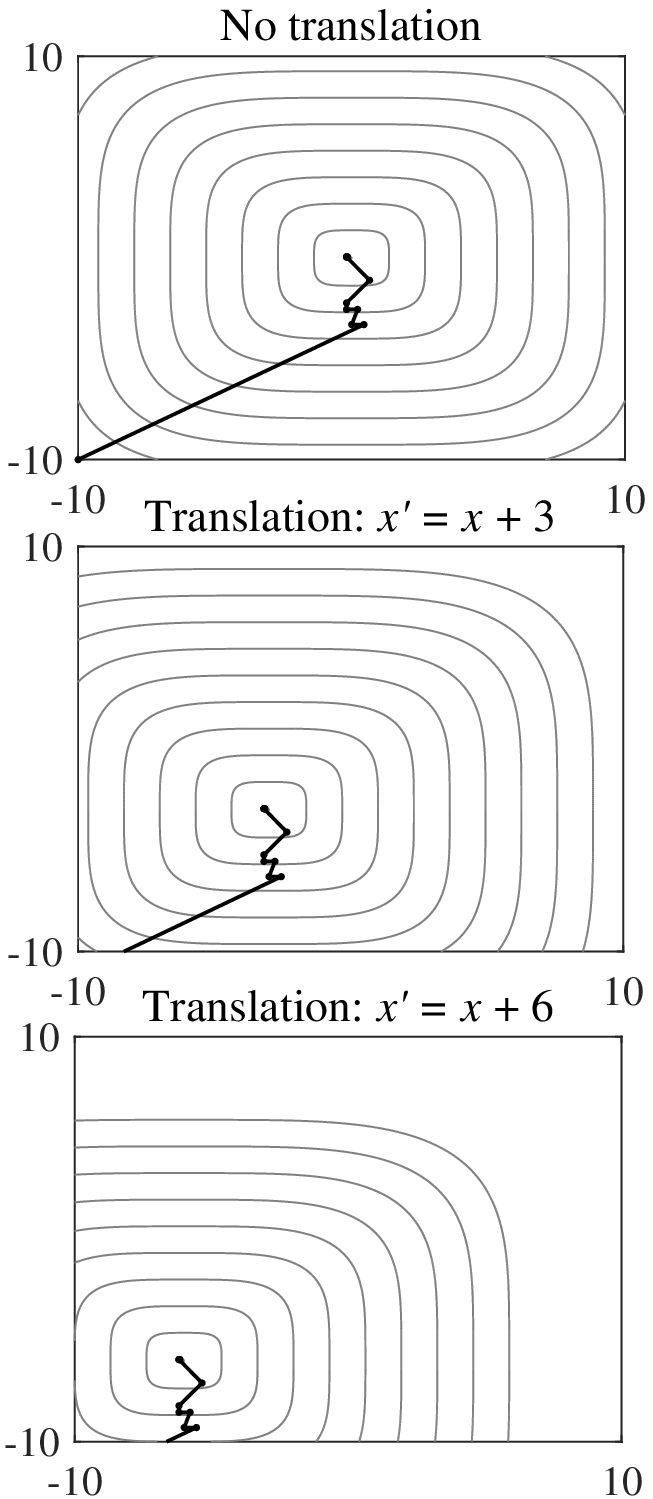}}\hfil
  \subfloat[SBX$^\prime$ based GA]{\includegraphics[width=0.5\linewidth]{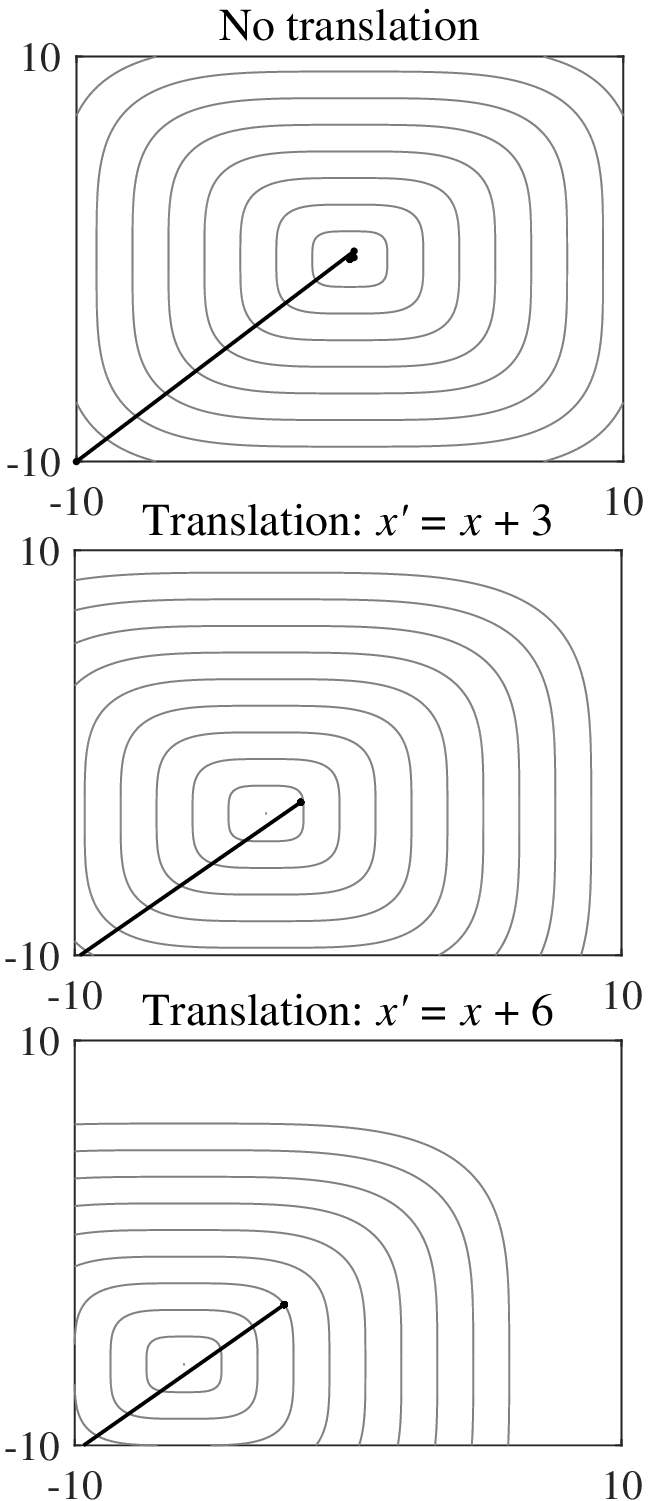}}
  \caption{Convergence profiles of SBX and SBX$^\prime$ based GAs with fixed random number seed on $f(x_1,x_2)=x_1^4+x_2^4$ with and without translation.}
  \label{fig:invariance1}
\end{figure}

To better illustrate this fact, Fig.~\ref{fig:invariance1} depicts the convergence profiles of SBX and SBX$^\prime$ based GAs on problem $f(x_1,x_2)=x_1^4+x_2^4$ with and without translation, where the random number seed is fixed and both the population size and the number of generations are set to 10. It can be found that the SBX based GA holds the same search behavior and can always converge to the global optimums of the three problems, whereas the SBX$^\prime$ based GA always converges to the origin. In short, the SBX operator is translation invariant but SBX$^\prime$ is not. Furthermore, Table~\ref{tab:example1} lists the mean and standard deviation of the minimum objective values obtained by SBX and SBX$^\prime$ based GAs on six benchmark problems \cite{Su2020Non}, averaged over 30 runs. The six problems have the same global optimum $(0,0,\dots,0)$, while the global optimum is changed to $(-6,-6,\dots,-6)$ if the problems are translated by $x^\prime=x+6$. For the original problems, the SBX$^\prime$ based GA significantly outperforms the SBX based GA since the former can quickly converge to the origin. While for the translated problems, the performance of SBX$^\prime$ based GA deteriorates considerably since it cannot converge to the translated global optimum; by contrast, the performance of SBX based GA keeps unchanged since the SBX operator is translation invariant.

\begin{table}[!t]
\footnotesize
\renewcommand{\arraystretch}{1.3}
\centering
\caption{Minimum Objective Values Obtained by SBX and SBX$^\prime$ Based GAs on Six Problems with and without Translation.}
\label{tab:example1}
\setlength{\tabcolsep}{0.5mm}{
\begin{tabular}{c|c|c}
\hline
Original problem&SBX based GA&\MR2{SBX$^\prime$ based GA}\\
($x_1,\dots,x_{30}\in[-10,10]$)&(translation invariant)\\
\hline
Schwefel's Function 2.22&5.4310e+0 (2.13e+0)&\hl{1.0965e-49 (1.29e-49)}\\
Schwefel's Function 2.21&4.5792e+0 (1.21e+0)&\hl{8.3896e-49 (1.23e-48)}\\
Quaric Function&1.6425e+3 (1.03e+3)&\hl{2.9477e-4 (2.66e-4)}\\
Griewank Function&4.7700e-1 (1.87e-1)&\hl{0.0000e+0 (0.00e+0)}\\
Ackley's Function&3.3728e+0 (3.75e-1)&\hl{8.8818e-16 (0.00e+0)}\\
Rastrigin's Function&4.2269e+1 (1.08e+1)&\hl{0.0000e+0 (0.00e+0)}\\
\hline\hline
Translated problem&SBX based GA&\MR2{SBX$^\prime$ based GA}\\
($x^\prime=x+6$)&(translation invariant)\\
\hline
Schwefel's Function 2.22&\hl{5.2237e+0 (1.88e+0)}&1.0083e+12 (2.18e+12)\\
Schwefel's Function 2.21&\hl{4.9401e+0 (1.24e+0)}&5.0000e+0 (0.00e+0)\\
Quaric Function&\hl{1.9875e+3 (1.62e+3)}&2.3718e+5 (1.53e+4)\\
Griewank Function&\hl{4.6346e-1 (1.17e-1)}&1.1523e+0 (4.76e-3)\\
Ackley's Function&\hl{3.5154e+0 (7.18e-1)}&1.2639e+1 (5.98e-4)\\
Rastrigin's Function&\hl{4.5695e+1 (1.01e+1)}&7.4895e+2 (1.83e-1)\\
\hline
\end{tabular}}
\end{table}

\subsection{Effects of Scale Invariance}

Similarly, a scaling of the search space can be regarded as a multiplication of each decision variable, i.e., $\mathcal{T}(x)=ax$, hence the scale invariance property can be defined as follows:
\begin{definition}[\textbf{Scale invariance}]
A variation operator $h(x_{1d},x_{2d},\dots)$ is scale invariant if and only if
\begin{equation}
h(ax_{1d},ax_{2d},\dots)=a\cdot h(x_{1d},x_{2d},\dots)
\end{equation}
holds for any real constant $a$.
\end{definition}

It can be found that the operators of SBX, DE, and CMA-ES are scale invariant but the operator of FEP is not. Considering that the step-size $\sigma$ is proportionally related to the decision space, the operator of CMA-ES described in (\ref{equ:CMAES}) can be rewritten as
\begin{equation}
h_{cmaes}(x_{md},l_d,u_d)=x_{md}+(u_d-l_d)\cdot\mathcal{N}(0,{\sigma^\prime_d}^2c)\ ,
\label{equ:example1}
\end{equation}
where $\sigma^\prime_d$ is a parameter within $(0,1]$ and $c$ is related to the covariance matrix $C$. Therefore,
\begin{equation}
\begin{aligned}
&h_{cmaes}(ax_{md},al_d,au_d)\\
&\qquad=ax_{md}+(au_d-al_d)\cdot\mathcal{N}(0,{\sigma^\prime_d}^2c)\\
&\qquad=a\cdot h_{cmaes}(x_{md},l_d,u_d)\ ,
\end{aligned}
\end{equation}
which means that the operator is scale invariant. On the other hand, the operator of FEP described in (\ref{equ:FEP}) can be rewritten as
\begin{equation}
h_{fep}(x_{d})=x_{d}+\mathcal{N}(0,\eta_d^2)\ ,
\end{equation}
hence
\begin{equation}
h_{fep}(ax_{d})=ax_{d}+\mathcal{N}(0,\eta_d^2)\neq a\cdot h_{fep}(x_{d})\ ,
\end{equation}
which means that the operator is not scale invariant.

\begin{figure}[!t]
  \centering
  \subfloat[CMA-ES (scale invariant)]{\includegraphics[width=0.5\linewidth]{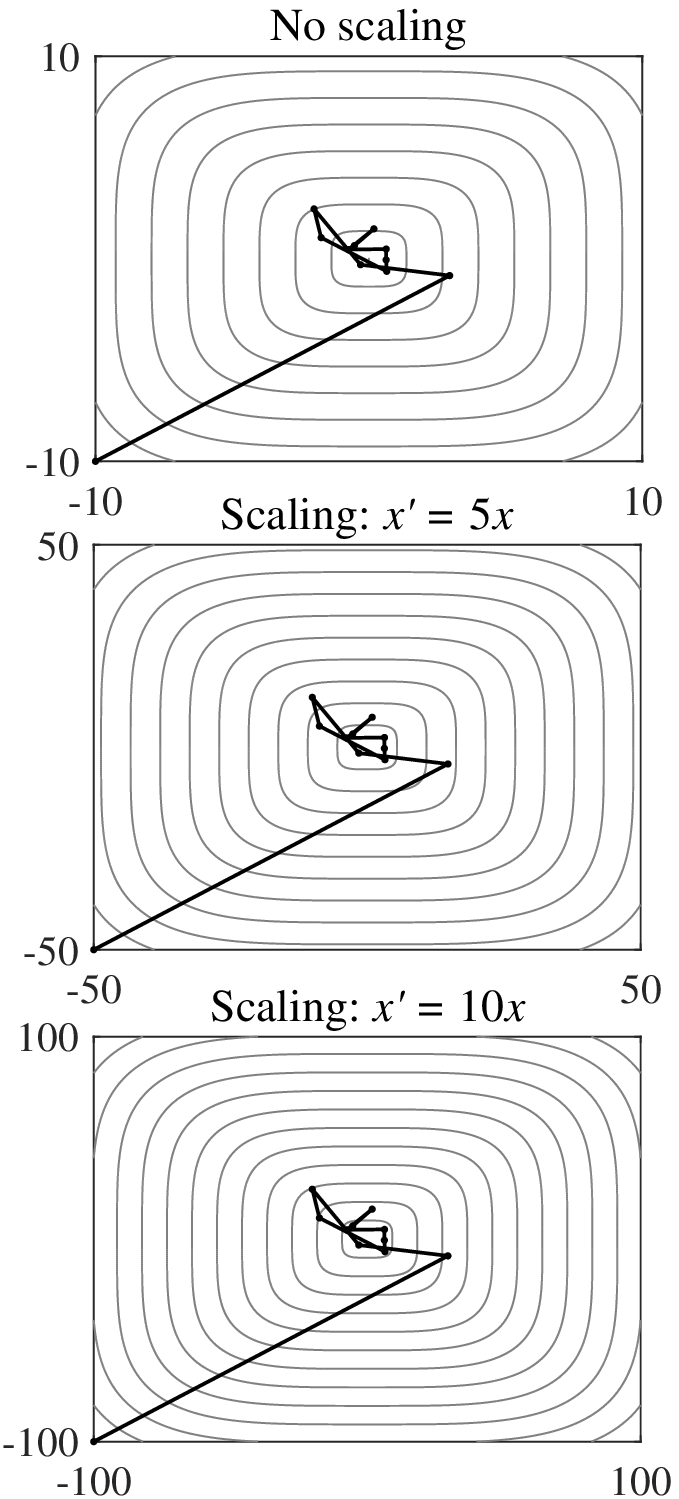}}\hfil
  \subfloat[FEP]{\includegraphics[width=0.5\linewidth]{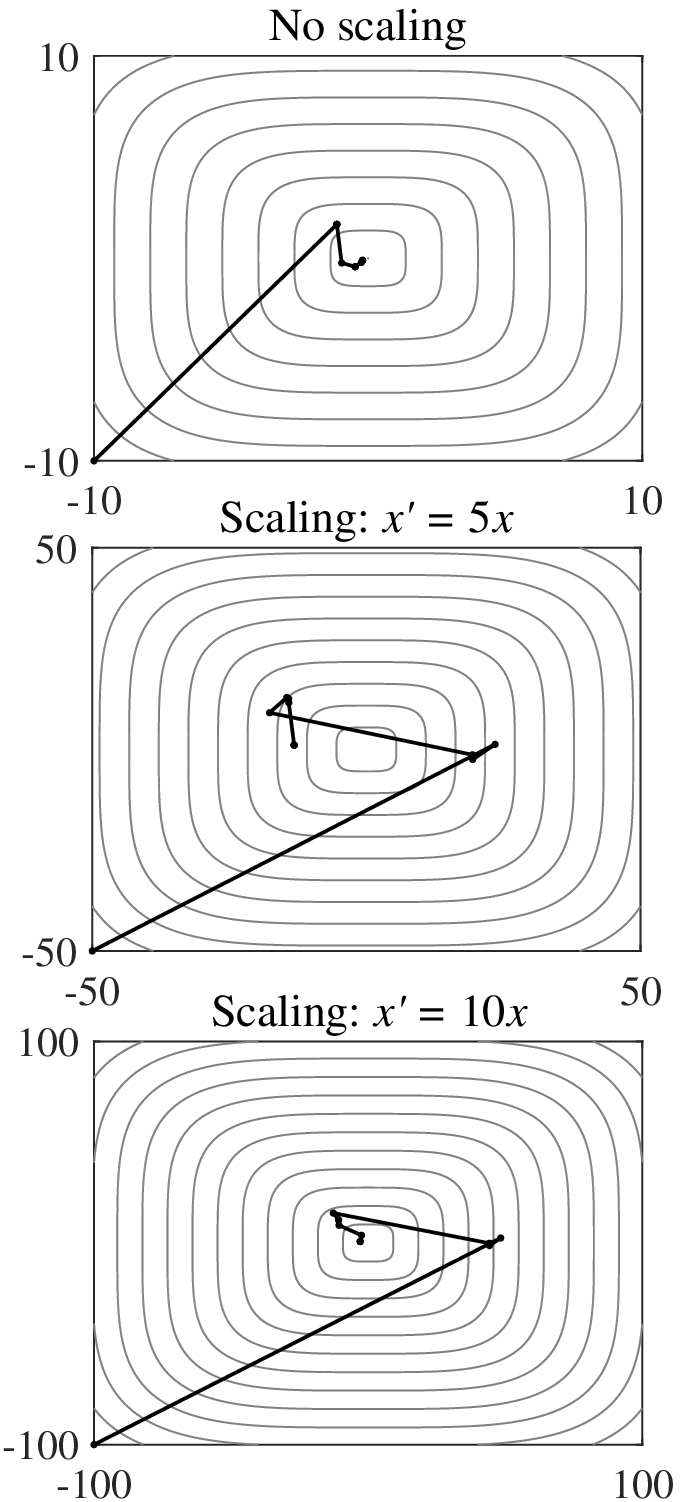}}
  \caption{Convergence profiles of CMA-ES and FEP with fixed random number seed on $f(x_1,x_2)=x_1^4+x_2^4$ with and without scaling.}
  \label{fig:invariance2}
\end{figure}

Fig.~\ref{fig:invariance2} plots the convergence profiles of CMA-ES and FEP on $f(x_1,x_2)=x_1^4+x_2^4$ with and without scaling. It can be seen that the search behaviors of CMA-ES are the same on the three problems with different scales, whereas the search behaviors of FEP are quite different. Moreover, Table~\ref{tab:example2} presents the mean and standard deviation of the minimum objective values obtained by CMA-ES and FEP on six benchmark problems. Obviously, CMA-ES is competitive to FEP on the original problems, while CMA-ES dramatically outperforms FEP on the scaled problems. This is because the operator of CMA-ES is scale invariant and thus exhibits the same performance on a problem with different scales; by contrast, the operator of FEP is not scale invariant and thus is sensitive to the scales of problems.

\begin{table}[!t]
\footnotesize
\renewcommand{\arraystretch}{1.3}
\centering
\caption{Minimum Objective Values Obtained by CMA-ES and FEP on Six Problems with and without Scaling.}
\label{tab:example2}
\setlength{\tabcolsep}{0.8mm}{
\begin{tabular}{c|c|c}
\hline
Original problem&CMA-ES&\MR2{FEP}\\
($x_1,\dots,x_{30}\in[-10,10]$)&(scale invariant)\\
\hline
Schwefel's Function 2.22&1.8226e+1 (3.38e+0)&\hl{1.6919e+1 (5.30e+0)}\\
Schwefel's Function 2.21&1.0000e+1 (0.00e+0)&\hl{4.6087e+0 (2.29e-1)}\\
Quaric Function&\hl{2.5024e+1 (9.92e+0)}&1.4005e+3 (1.69e+3)\\
Griewank Function&\hl{3.2242e-1 (5.74e-2)}&8.5491e-1 (1.18e-1)\\
Ackley's Function&\hl{2.7634e+0 (1.71e-1)}&5.7111e+0 (5.97e-1)\\
Rastrigin's Function&2.1628e+2 (4.87e+0)&\hl{1.6904e+2 (2.52e+1)}\\
\hline\hline
Scaled problem&CMA-ES&\MR2{FEP}\\
($x^\prime=10x$)&(scale invariant)\\
\hline
Schwefel's Function 2.22&\hl{1.9390e+1 (2.70e+0)}&4.6714e+10 (5.47e+10)\\
Schwefel's Function 2.21&\hl{1.3107e+0 (1.36e-1)}&8.4873e+0 (5.56e-1)\\
Quaric Function&\hl{2.6711e+1 (2.22e+1)}&3.8536e+5 (2.35e+4)\\
Griewank Function&\hl{2.8246e-1 (9.96e-2)}&1.1561e+0 (8.63e-3)\\
Ackley's Function&\hl{3.0355e+0 (3.42e-1)}&1.3471e+1 (5.21e-1)\\
Rastrigin's Function&\hl{2.1651e+2 (1.40e+1)}&9.2797e+2 (3.64e+1)\\
\hline
\end{tabular}}
\end{table}

\subsection{Effects of Rotation Invariance}

A rotation of the search space can be regarded as a matrix multiplication of each decision vector, i.e., $\mathcal{T}(\mathbf{x})=\mathbf{x}M$, hence the rotation invariance property can be defined as follows:
\begin{definition}[\textbf{Rotation invariance}]
A variation operator $h(\mathbf{x}_1,\mathbf{x}_2,\dots)$ is rotation invariant if and only if
\begin{equation}
h(\mathbf{x}_1M,\mathbf{x}_2M,\dots)=h(\mathbf{x}_1,\mathbf{x}_2,\dots)M
\end{equation}
holds for any orthogonal matrix $M$.
\end{definition}

Note that here the decision variables on all dimensions $\mathbf{x}_1,\mathbf{x}_2,\dots$ rather than those on a single dimension $x_{1d},x_{2d},\dots$ should be considered. It can be deduced that the mutation operator of DE is rotation invariant while the operators of SBX, CMA-ES, and FEP are not. In particular, the mutation operator of DE described in (\ref{equ:DE}) can be rewritten as
\begin{equation}
h_{de}(\mathbf{x}_1,\mathbf{x}_2,\mathbf{x}_3)=\mathbf{x}_1+F\cdot(\mathbf{x}_2-\mathbf{x}_3)\ ,
\end{equation}
hence
\begin{equation}
\begin{aligned}
h_{de}(\mathbf{x}_1M,\mathbf{x}_2M,\mathbf{x}_3M)=&\ \mathbf{x}_1M+F\cdot(\mathbf{x}_2M-\mathbf{x}_3M)\\ =&\ h_{de}(\mathbf{x}_1,\mathbf{x}_2,\mathbf{x}_3)M\ ,
\end{aligned}
\end{equation}
and the operator is rotation invariant. By contrast, since the SBX operator described in (\ref{equ:SBX2}) can be rewritten as
\begin{equation}
h_{sbx}(\mathbf{x}_1,\mathbf{x}_2)=\mathbf{x}_1+(\mathbf{x}_2-\mathbf{x}_1)B\ ,
\end{equation}
where
\begin{equation}
\footnotesize
B=\left(\begin{array}{cccc}
0.5(1\pm\beta_1)&0&\cdots&0\\
0&0.5(1\pm\beta_2)&\cdots&0\\
\cdots&\cdots&\cdots&\cdots\\
0&0&0&0.5(1\pm\beta_D)\\
\end{array}\right)\normalsize\ .
\end{equation}
Therefore,
\begin{equation}
h_{sbx}(\mathbf{x}_1M,\mathbf{x}_2M)=\mathbf{x}_1M+(\mathbf{x}_2-\mathbf{x}_1)MB
\end{equation}
and
\begin{equation}
h_{sbx}(\mathbf{x}_1,\mathbf{x}_2)M=\mathbf{x}_1M+(\mathbf{x}_2-\mathbf{x}_1)BM\ .
\end{equation}
That is, $h_{sbx}(\mathbf{x}_1M,\mathbf{x}_2M)=h_{sbx}(\mathbf{x}_1,\mathbf{x}_2)M$ holds only if $MB=BM$, i.e., $\beta_1=\beta_2=\dots=\beta_D$, which is almost impossible since they are independent random numbers. Hence, the SBX operator is not rotation invariant.

\begin{figure}[!t]
  \centering
  \subfloat[DE (rotation invariant)]{\includegraphics[width=0.5\linewidth]{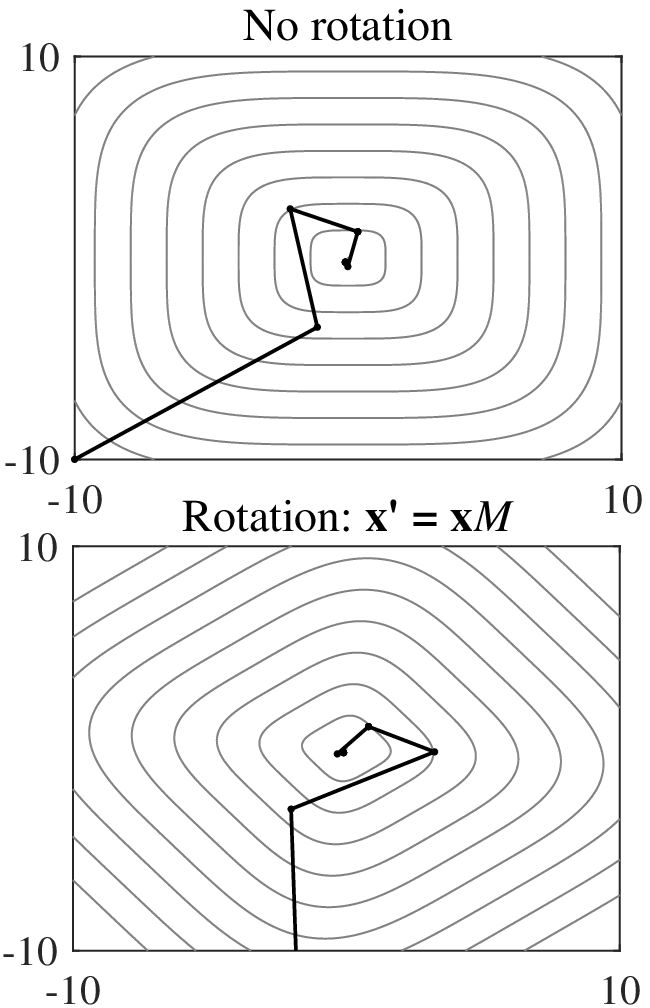}}\hfil
  \subfloat[SBX based GA]{\includegraphics[width=0.5\linewidth]{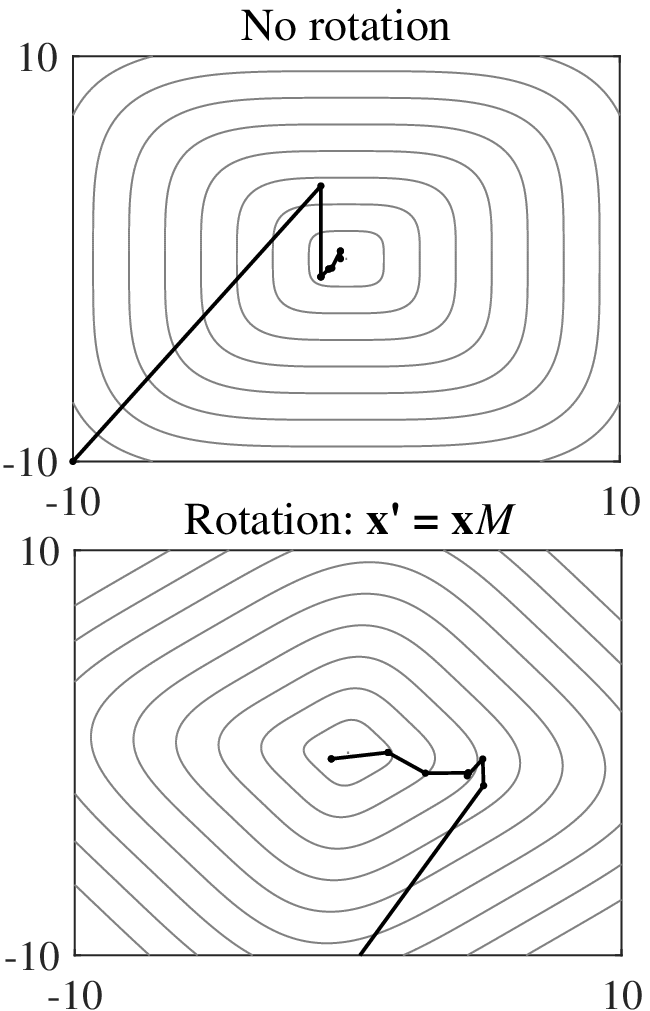}}
  \caption{Convergence profiles of DE and SBX based GA with fixed random number seed on $f(x_1,x_2)=x_1^4+x_2^4$ with and without rotation, where $M$ is a randomly generated orthogonal matrix.}
  \label{fig:invariance3}
\end{figure}

Fig.~\ref{fig:invariance3} shows the convergence profiles of the mutation based DE and SBX based GA on $f(x_1,x_2)=x_1^4+x_2^4$ with and without rotation. As can be seen, the search behavior of DE keeps unchanged on the two problems, where the convergence profile is rotated together with the search space. On the contrary, the search behavior of GA is unstable and the convergence profiles are different on the two problems. As can be further observed from Table~\ref{tab:example3}, the performance of DE is similar to and much better than GA on six benchmark problems with and without rotation, respectively, which is consistent with the facts that the mutation operator of DE is rotation invariant while the SBX operator is not. In fact, the superiority of DE on problems with complicated variable linkages \cite{MOEA/D-DE} is mainly due to its rotation invariance property.

\begin{table}[!t]
\footnotesize
\renewcommand{\arraystretch}{1.3}
\centering
\caption{Minimum Objective Values Obtained by DE and SBX Based GA on Six Problems with and without Rotation.}
\label{tab:example3}
\setlength{\tabcolsep}{0.8mm}{
\begin{tabular}{c|c|c}
\hline
Original problem&DE&\MR2{SBX based GA}\\
($x_1,\dots,x_{30}\in[-10,10]$)&(rotation invariant)\\
\hline
Schwefel's Function 2.22&1.9799e+1 (2.08e+0)&\hl{5.4310e+0 (2.13e+0)}\\
Schwefel's Function 2.21&\hl{1.5589e+0 (1.87e-1)}&4.5792e+0 (1.21e+0)\\
Quaric Function&\hl{1.8303e+2 (9.67e+1)}&1.6425e+3 (1.03e+3)\\
Griewank Function&4.9032e-1 (1.01e-1)&\hl{4.7700e-1 (1.87e-1)}\\
Ackley's Function&3.7861e+0 (2.96e-1)&\hl{3.3728e+0 (3.75e-1)}\\
Rastrigin's Function&2.4783e+2 (2.34e+1)&\hl{4.2269e+1 (1.08e+1)}\\
\hline\hline
Rotated problem&DE&\MR2{SBX based GA}\\
($\mathbf{x}^\prime=\mathbf{x}M$)&(rotation invariant)\\
\hline
Schwefel's Function 2.22&\hl{1.7401e+1 (3.38e+0)}&2.8069e+1 (4.32e+0)\\
Schwefel's Function 2.21&\hl{1.5535e+0 (2.21e-1)}&2.0725e+0 (4.82e-1)\\
Quaric Function&\hl{1.2403e+2 (7.80e+1)}&9.8479e+2 (1.24e+3)\\
Griewank Function&\hl{4.6714e-1 (7.22e-2)}&4.9478e-1 (1.20e-1)\\
Ackley's Function&\hl{3.8477e+0 (2.24e-1)}&4.0805e+0 (5.49e-1)\\
Rastrigin's Function&\hl{2.4588e+2 (1.23e+1)}&2.5095e+2 (1.12e+1)\\
\hline
\end{tabular}}
\end{table}

\section{Sufficient and Necessary Condition of Achieving Invariance Properties}

It can be concluded from the above analysis that the three invariance properties are critical to the robustness of operators, and we can judge whether an operator possesses these properties according to their definitions. However, it is still tricky to consider them in the design of new operators explicitly. Therefore, this section deduces the generic form of translation, scale, and rotation invariant operators by three steps.

\subsection{Theoretical Derivation}

Firstly, let $\mathbf{w}=(w_1,w_2,\dots)=(x_{1d},x_{2d},\dots)$, then a translation invariant operator should satisfy
\begin{equation}
h(\mathbf{w}+b)=h(\mathbf{w})+b\ .
\label{equ:deduce1}
\end{equation}
To find the generic form of $h(\mathbf{w})$, it is first expanded in a first order Taylor series at an arbitrary point $\mathbf{w}_0$:
\begin{equation}
h(\mathbf{w})=h(\mathbf{w}_0)+(\mathbf{w}-\mathbf{w}_0)\nabla h(\mathbf{w}_0^\prime)\ ,
\end{equation}
where $\nabla=(\frac{\partial}{\partial w_1},\frac{\partial}{\partial w_2},\dots)^T$ and $\mathbf{w}_0^\prime$ is an unknown point. Based on (\ref{equ:deduce1}), we have
\begin{equation}
\begin{aligned}
&h(\mathbf{w}_0)+(\mathbf{w}+b-\mathbf{w}_0)\nabla h(\mathbf{w}_0^\prime)\\
&\qquad=h(\mathbf{w}_0)+(\mathbf{w}-\mathbf{w}_0)\nabla h(\mathbf{w}_0^{\prime\prime})+b\ ,
\end{aligned}
\label{equ:deduce2}
\end{equation}
where $\mathbf{w}_0^\prime$ and $\mathbf{w}_0^{\prime\prime}$ are unknown points determined by $\mathbf{w}_0$. Since (\ref{equ:deduce2}) holds for any $b$, we only consider the components including $b$ in (\ref{equ:deduce2}):
\begin{equation}
bI\nabla h(\mathbf{w}_0^\prime)=b\ ,
\end{equation}
where $I$ is a vector of ones. That is,
\begin{equation}
I\nabla h(\mathbf{w}_0^\prime)=1
\end{equation}
holds for any $\mathbf{w}_0^\prime$, which measn that
\begin{equation}
\frac{\partial h}{\partial w_1}+\frac{\partial h}{\partial w_2}+\cdots=1\ .
\label{equ:deduce3}
\end{equation}
Obviously, (\ref{equ:deduce3}) is a quasilinear first-order nonhomogeneous partial differential equation \cite{PDE}. Let $h=h(\mathbf{w})$ be the solution of (\ref{equ:deduce3}) determined by a function $g(\mathbf{w},h)=0$, then the following homogeneous partial differential equation can be obtained:
\begin{equation}
\frac{\partial g}{\partial h}+\frac{\partial g}{\partial w_1}+\frac{\partial g}{\partial w_2}+\cdots=0\ ,
\label{equ:deduce4}
\end{equation}
and thus the following first integrals can be obtained:
\begin{equation}
\left\{
\begin{aligned}
h-w_1=&\ c_1\\
w_1-w_2=&\ c_2\\
w_2-w_3=&\ c_3\\
\cdots\ \ \cdots
\end{aligned}
\right.,
\end{equation}
Therefore, the solution of (\ref{equ:deduce4}) is
\begin{equation}
g=g(h-w_1,w_1-w_2,w_2-w_3,\dots)\ .
\end{equation}
Since $g=0$, there must exists a function $\psi$ such that
\begin{equation}
h-w_1=\psi(w_1-w_2,w_2-w_3,\dots)\ ,
\end{equation}
which is equivalent to
\begin{equation}
h(\mathbf{w})=w_1+\psi(w_2-w_1,w_3-w_2,\dots)\ .
\label{equ:deduce5}
\end{equation}
Moreover, it is obvious that (\ref{equ:deduce5}) satisfies (\ref{equ:deduce1}), hence (\ref{equ:deduce5}) is a sufficient and necessary condition of (\ref{equ:deduce1}), and the following theorem can be given:
\begin{theorem}[\textbf{Translation invariant operator}]
A continuously differentiable variation operator $h(x_{1d},x_{2d},\dots)$ is translation invariant if and only if it has the following form:
\begin{equation}
h(x_{1d},x_{2d},\dots)=x_{1d}+\psi(x_{2d}-x_{1d},x_{3d}-x_{2d},\dots)\ ,
\end{equation}
where $\psi$ can be any continuously differentiable function.
\end{theorem}

Secondly, a scale invariant operator should satisfy
\begin{equation}
h(a\mathbf{w})=a\cdot h(\mathbf{w})\ ,
\label{equ:deduce9}
\end{equation}
according to (\ref{equ:deduce5}), a translation and scale invariant operator should satisfy
\begin{equation}
\begin{aligned}
&aw_1+\psi(aw_2-aw_1,aw_3-aw_2,\dots)\\
&\qquad=aw_1+a\cdot\psi(aw_2-aw_1,aw_3-aw_2,\dots)\ .
\end{aligned}
\end{equation}
Let $\mathbf{v}=(v_1,v_2,\dots)=(w_2-w_1,w_3-w_2,\dots)$, we have
\begin{equation}
\psi(a\mathbf{v})=a\cdot \psi(\mathbf{v}),
\end{equation}
and the first order Taylor series expansion at an arbitrary point $\mathbf{v}_0$ is
\begin{equation}
\begin{aligned}
&\psi(\mathbf{v}_0)+(a\mathbf{v}-\mathbf{v}_0)\nabla \psi(\mathbf{v}_0^\prime)\\
&\qquad=a\psi(\mathbf{v}_0)+a(\mathbf{v}-\mathbf{v}_0)\nabla \psi(\mathbf{v}_0^{\prime\prime})\ ,
\end{aligned}
\label{equ:deduce6}
\end{equation}
where $\mathbf{v}_0^\prime$ and $\mathbf{v}_0^{\prime\prime}$ are unknown points determined by $\mathbf{v}_0$. Since (\ref{equ:deduce6}) holds for any $a$, we only consider the components including $a$ in (\ref{equ:deduce6}), that is,
\begin{equation}
\psi(\mathbf{v}_0)=\mathbf{v}_0\nabla \psi(\mathbf{v}_0^\prime)
\end{equation}
must hold for any $\mathbf{v}_0$ and $\mathbf{v}_0^\prime$, which means that
\begin{equation}
v_1\frac{\partial \psi}{\partial v_1}+v_2\frac{\partial \psi}{\partial v_2}+\cdots=\psi\ .
\end{equation}
Let $g(\mathbf{v},\psi)=0$, the following homogeneous partial differential equation can be obtained:
\begin{equation}
\psi\frac{\partial g}{\partial \psi}+v_1\frac{\partial g}{\partial v_1}+v_2\frac{\partial g}{\partial v_2}+\cdots=0\ ,
\label{equ:deduce7}
\end{equation}
and thus the following first integrals can be obtained:
\begin{equation}
\left\{
\begin{aligned}
\ln \psi-\ln v_1=&\ c_1\\
\ln v_1-\ln v_2=&\ c_2\\
\ln v_2-\ln v_3=&\ c_3\\
\cdots\ \ \cdots
\end{aligned}
\right.,
\end{equation}
Therefore, the solution of (\ref{equ:deduce7}) is
\begin{equation}
g=g(\ln \psi-\ln v_1,\ln v_1-\ln v_2,\ln v_2-\ln v_3,\dots)\ .
\end{equation}
Since $g=0$, there must exists a function $\varphi$ such that
\begin{equation}
\ln \psi-\ln v_1=\varphi(\ln v_1-\ln v_2,\ln v_2-\ln v_3,\dots)\ ,
\end{equation}
hence the generic form of $\psi(\mathbf{v})$ can be determined:
\begin{equation}
\ln \psi(\mathbf{v})=\ln v_1 + \varphi\left(\ln \frac{v_1}{v_2},\ln \frac{v_2}{v_3},\dots\right)\ .
\end{equation}
Let $\varphi(u_1,u_2,\dots)=\ln \phi(e^\frac{1}{u_1},e^\frac{1}{u_2},\dots)$, then
\begin{equation}
\psi(\mathbf{v})=v_1\phi\left(\frac{v_2}{v_1},\frac{v_3}{v_2},\dots\right)\ .
\end{equation}
According to (\ref{equ:deduce5}), we have
\begin{equation}
h(\mathbf{w})=w_1+(w_2-w_1)\phi\left(\frac{w_3-w_2}{w_2-w_1},\frac{w_4-w_3}{w_3-w_2},\dots\right).
\label{equ:deduce8}
\end{equation}
Moreover, it is obvious that (\ref{equ:deduce8}) satisfies both (\ref{equ:deduce1}) and (\ref{equ:deduce9}), hence (\ref{equ:deduce8}) is a sufficient and necessary condition of (\ref{equ:deduce1}) and (\ref{equ:deduce9}), and the following theorem can be given:
\begin{theorem}[\textbf{Translation and scale invariant operator}]
A continuously differentiable variation operator $h(x_{1d},x_{2d},\dots)$ is translation and scale invariant if and only if it has the following form:
\begin{equation}
\footnotesize
h(x_{1d},x_{2d},\dots)=x_{1d}+(x_{2d}-x_{1d})\phi\left(\frac{x_{3d}-x_{2d}}{x_{2d}-x_{1d}},\frac{x_{4d}-x_{3d}}{x_{3d}-x_{2d}},\dots\right),
\end{equation}
where $\phi$ can be any continuously differentiable function.
\end{theorem}

Thirdly, a rotation invariant operator should satisfy
\begin{equation}
\footnotesize
h\left(\sum_{i=1}^{D}m_{id}x_{1i},\sum_{i=1}^{D}m_{id}x_{2i},\dots\right)=\sum_{i=1}^{D}m_{id}\cdot h(x_{1i},x_{2i},\dots)\ ,
\label{equ:deduce10}
\end{equation}
where $m_{id}\in M$ and $d=1,\dots,D$. Let $\phi(\mathbf{w})=\varphi(w_1,\prod_{i=1}^{2}w_i,\prod_{i=1}^{3}w_i,\dots)$, then (\ref{equ:deduce8}) is equivalent to
\begin{equation}
h(\mathbf{w})=w_1+(w_2-w_1)\varphi\left(\frac{w_3-w_2}{w_2-w_1},\frac{w_4-w_3}{w_2-w_1},\dots\right)\ .
\label{equ:deduce12}
\end{equation}
According to (\ref{equ:deduce10}), a translation, scale, and rotation invariant operator should satisfy
\begin{equation}
\begin{aligned}
&\sum_{i=1}^{D}m_{id}x_{1i}+\left(\sum_{i=1}^{D}m_{id}x_{2i}-\sum_{i=1}^{D}m_{id}x_{1i}\right)\\
&\cdot\varphi\left(\frac{\sum_{i=1}^{D}m_{id}x_{3i}-\sum_{i=1}^{D}m_{id}x_{2i}}{\sum_{i=1}^{D}m_{id}x_{2i}-\sum_{i=1}^{D}m_{id}x_{1i}},\dots\right)\\
&=\sum_{i=1}^{D}m_{id}\left[x_{1i}+(x_{2i}-x_{1i})\varphi\left(\frac{x_{3i}-x_{2i}}{x_{2i}-x_{1i}},\dots\right)\right]\ ,
\end{aligned}
\end{equation}
which is equivalent to
\begin{equation}
\begin{aligned}
&\sum_{i=1}^{D}m_{id}(x_{2i}-x_{1i})\cdot\varphi\left(\frac{\sum_{i=1}^{D}m_{id}(x_{3i}-x_{2i})}{\sum_{i=1}^{D}m_{id}(x_{2i}-x_{1i})},\dots\right)\\
&=\sum_{i=1}^{D}m_{id}(x_{2i}-x_{1i})\varphi\left(\frac{m_{id}(x_{3i}-x_{2i})}{m_{id}(x_{2i}-x_{1i})},\dots\right)\ .
\end{aligned}
\end{equation}
Let $\mathbf{u}_i=(u_{i1},u_{i2},\dots)=(m_{id}(x_{2i}-x_{1i}),m_{id}(x_{3i}-x_{2i}),\dots)$, we have
\begin{equation}
\sum_{i=1}^{D}u_{i1}\cdot\varphi\left(\frac{\sum_{i=1}^{D}u_{i2}}{\sum_{i=1}^{D}u_{i1}},\dots\right)=\sum_{i=1}^{D}u_{i1}\varphi\left(\frac{u_{i2}}{u_{i1}},\dots\right)\ ,
\end{equation}
and the first order Taylor series expansion at an arbitrary point $\mathbf{u}_0=(u_{01},u_{02},\dots)$ is
\begin{equation}
\begin{aligned}
&\sum_{i=1}^{D}u_{i1}\cdot\left[\varphi(\mathbf{u}_0)+\left(\frac{\sum_{i=1}^{D}u_{i2}}{\sum_{i=1}^{D}u_{i1}}-u_{01},\dots\right)\nabla \varphi(\mathbf{u}_0^\prime)\right]\\
&=u_{11}\cdot\left[\varphi(\mathbf{u}_0)+\left(\frac{u_{12}}{u_{11}}-u_{01},\dots\right)\nabla \varphi(\mathbf{u}_0^{\prime\prime})\right]\\
&+u_{21}\cdot\left[\varphi(\mathbf{u}_0)+\left(\frac{u_{22}}{u_{21}}-u_{01},\dots\right)\nabla \varphi(\mathbf{u}_0^{\prime\prime\prime})\right]\\
&+\cdots\ ,
\end{aligned}
\end{equation}
which can be simplified as
\begin{equation}
\begin{aligned}
&\left(\sum_{i=1}^{D}u_{i2}-u_{01}\sum_{i=1}^{D}u_{i1},\dots\right)\nabla \varphi(\mathbf{u}_0^\prime)\qquad\\
&\qquad=\left(u_{12}-u_{01}u_{11},\dots\right)\nabla \varphi(\mathbf{u}_0^{\prime\prime})\\ &\qquad+\left(u_{22}-u_{01}u_{21},\dots\right)\nabla \varphi(\mathbf{u}_0^{\prime\prime\prime})\\
&\qquad+\cdots\ ,
\end{aligned}
\label{equ:deduce11}
\end{equation}
where $\mathbf{u}_0^\prime,\mathbf{u}_0^{\prime\prime},\dots$ are unknown points determined by $\mathbf{u}_0$. Since (\ref{equ:deduce11}) holds for any $u_{12},u_{22},\dots$, we have
\begin{equation}
\nabla\varphi(\mathbf{u}_0^\prime)=\nabla\varphi(\mathbf{u}_0^{\prime\prime})=\nabla\varphi(\mathbf{u}_0^{\prime\prime\prime})=\cdots=\mathbf{c}\ ,
\end{equation}
and the form of $\varphi(\mathbf{u})$ can only be
\begin{equation}
\varphi(\mathbf{u})=c_0+c_1u_1+c_2u_2+\cdots\ ,
\end{equation}
where $c_0,c_1,c_2,\dots$ are constants. According to (\ref{equ:deduce12}),
\begin{equation}
h(\mathbf{w})=w_1+c_0(w_2-w_1)+c_1(w_3-w_2)+c_2(w_4-w_3)+\cdots\ .
\end{equation}
Let
\begin{equation}
\left\{
\begin{aligned}
&\ 1-c_0=r_1\\
&c_0-c_1=r_2\\
&c_1-c_2=r_3\\
&\cdots\ \ \cdots\\
\end{aligned}
\right.\ ,
\end{equation}
we have
\begin{equation}
h(\mathbf{w})=r_1w_1+r_2w_2+r_3w_3+\cdots
\label{equ:deduce13}
\end{equation}
and $r_1+r_2+r_3+\cdots=1$. Moreover, it is obvious that (\ref{equ:deduce13}) satisfies (\ref{equ:deduce1}), (\ref{equ:deduce9}), and (\ref{equ:deduce10}), hence (\ref{equ:deduce13}) is a sufficient and necessary condition of (\ref{equ:deduce1}), (\ref{equ:deduce9}), and (\ref{equ:deduce10}), and the following theorem can be given:
\begin{theorem}[\textbf{Translation, scale, and rotation invariant operator}]
A continuously differentiable variation operator $h(x_{1d},x_{2d},\dots)$ is translation, scale, and rotation invariant if and only if it has the following form:
\begin{equation}
h(x_{1d},x_{2d},\dots)=r_1x_{1d}+r_2x_{2d}+r_3x_{3d}+\cdots\ ,
\label{equ:example2}
\end{equation}
where $r_1,r_2,r_3,\dots$ can be any real constants satisfying $r_1+r_2+r_3+\cdots=1$.
\end{theorem}

\subsection{Remarks}

According to the above theorem, the following three corollaries can be given:
\begin{enumerate}
\item An operator satisfying (\ref{equ:example2}) is scale invariant.
\item An operator satisfying (\ref{equ:example2}) with $r_1+r_2+r_3+\cdots=1$ is scale and translation invariant.
\item An operator satisfying (\ref{equ:example2}) with $r_1,r_2,r_3,\dots$ being constants is scale and rotation invariant.
\end{enumerate}
The theoretical proofs of these corollaries are given in the following.
\begin{corollary}
If a variation operator satisfying
\begin{equation}
h(x_{1d},x_{2d},\dots)=r_1x_{1d}+r_2x_{2d}+r_3x_{3d}+\cdots\ ,
\end{equation}
then it is scale invariant.
\end{corollary}

\begin{IEEEproof}
Since
\begin{equation}
\begin{aligned}
h(ax_{1d},ax_{2d},\dots)=&\ ar_1x_{1d}+ar_2x_{2d}+ar_3x_{3d}+\cdots\\
=&\ ah(x_{1d},x_{2d},\dots)\ ,
\end{aligned}
\end{equation}
the operator is scale invariant.
\end{IEEEproof}

\begin{corollary}
If a variation operator satisfying
\begin{equation}
h(x_{1d},x_{2d},\dots)=r_1x_{1d}+r_2x_{2d}+r_3x_{3d}+\cdots
\end{equation}
with $r_1+r_2+r_3+\cdots=1$, then it is scale and translation invariant.
\end{corollary}

\begin{IEEEproof}
Since
\begin{equation}
\begin{aligned}
&h(x_{1d}+b,x_{2d}+b,\dots)\\
&\quad=r_1(x_{1d}+b)+r_2(x_{2d}+b)+r_3(x_{3d}+b)+\cdots\\
&\quad=r_1x_{1d}+r_2x_{2d}+r_3x_{3d}+\cdots+(r_1+r_2+r_3+\cdots)b\\
&\quad=h(x_{1d},x_{2d},\dots)+b\ ,
\end{aligned}
\end{equation}
according to Corollary 1, the operator is scale and translation invariant.
\end{IEEEproof}

\begin{corollary}
If a variation operator satisfying
\begin{equation}
h(x_{1d},x_{2d},\dots)=r_1x_{1d}+r_2x_{2d}+r_3x_{3d}+\cdots
\end{equation}
with $r_1,r_2,r_3,\dots$ being constants, then it is scale and rotation invariant.
\end{corollary}

\begin{IEEEproof}
Since $r_1,r_2,r_3,\dots$ are constants, we have
\begin{equation}
h(\mathbf{x}_1,\mathbf{x}_2,\dots)=r_1\mathbf{x}_1+r_2\mathbf{x}_2+r_3\mathbf{x}_3+\cdots\ .
\end{equation}
Therefore,
\begin{equation}
\begin{aligned}
&h(\mathbf{x}_1M,\mathbf{x}_2M,\dots)\\
&\qquad=r_1(\mathbf{x}_1M)+r_2(\mathbf{x}_2M)+r_3(\mathbf{x}_3M)+\cdots\\
&\qquad=(r_1\mathbf{x}_1)M+(r_2\mathbf{x}_2)M+(r_3\mathbf{x}_3)M+\cdots\\
&\qquad=h(\mathbf{x}_1,\mathbf{x}_2,\dots)M\ ,
\end{aligned}
\end{equation}
according to Corollary 1, the operator is scale and rotation invariant.
\end{IEEEproof}
Generally, most existing operators including those in GA, DE, and CMA-ES meet the second corollary and are scale and translation invariant. However, the operators in GA and CMA-ES are not rotation invariant since the weights (i.e., $\beta$ in (\ref{equ:SBX}) and $\mathcal{N}(0,{\sigma^\prime_d}^2c)$ in (\ref{equ:example1})) vary on different dimensions. By contrast, the mutation operator of DE is rotation invariant since the weights (i.e., $F$ in (\ref{equ:DE})) keep unchanged on all dimensions.

It should be noted that a rotation invariant operator does not necessarily lead to a rotation invariant metaheuristic, and vice versa. For example, the mutation operator of DE is rotation variant, but DE is not rotation invariant due to the crossover operator with $CR<1$ \cite{Caraffini2019Study}. The operator of CMA-ES is not rotation invariant, but CMA-ES is rotation invariant due to the rotation angle adaptive covariance matrix $C$ \cite{algorithm-CMAES}. In fact, a rotation invariant operator may not obtain good performance since offspring solutions can only be the linear combinations of parents. In practice, the rotation invariance can be achieved by rotating offspring solutions according to a covariance matrix learnt from the population \cite{algorithm-CMAES,Pan2021Adaptive}.

\section{Principled Approach for Designing Operators}

The function of (\ref{equ:example2}) reveals the generic form of translation, scale, and rotation invariant operators, while the optimal values of the weights $r_1,r_2,r_3,\dots$ are not provided. Therefore, this section proposes a principled approach for designing new operators, which searches for high-performance operators by optimizing the weights.

\subsection{Parameterization of Variation Operators}

In order to introduce randomness, the proposed approach AutoV represents each weight by an independent normal distribution, and it optimizes the mean and variance of each normal distribution instead of the weight. Formally, the proposed AutoV searches for the operators having the following form:
\begin{equation}
\begin{aligned}
h(x_{1d},\dots,x_{td})=&\ \sum_{i=1}^{t}r_ix_{id}\\
{\rm s.\,t.}\ \ \sum_{i=1}^{t}r_i=&\ 1,\ r_{i}\sim\mathcal{N}(\mu_i,\sigma_i^2)
\end{aligned}
\ ,
\label{equ:AutoV1}
\end{equation}
where $\mu_i\in[-1,1]$ and $\sigma_i\in[0,1]$ are the parameters to be optimized. Note that the value of $r_i$ keeps unchanged for $d=1,\dots,D$, hence $r_i$ is still a constant for generating each offspring solution. In the following, we demonstrate that an algorithm equipped with the above operator can converge to the global optimum of continuous optimization problems if given sufficient time.
\begin{theorem}
Given a continuous function $f(\mathbf{x})$ and its global optimal solution $\mathbf{x}^\ast$, for any $\epsilon>0$ and a metaheuristic based on an elite strategy and a variation operator satisfying
\begin{equation}
\begin{aligned}
h(x_{1d},\dots,x_{td})=&\ (1-\sum_{i=2}^{t}r_i)x_{1d}+\sum_{i=2}^{t}r_ix_{id}\\
{\rm s.t.}\ \ r_{i+1}\sim&\ \mathcal{N}(\mu_i,\sigma_i^2),\ i=1,\dots,t-1
\end{aligned}
\ ,
\label{equ:AutoV1}
\end{equation}
we have
\begin{equation}
\lim_{g\rightarrow\infty}P(|f(\mathbf{x}^g)-f(\mathbf{x}^\ast)|<\epsilon)=1\ ,
\end{equation}
where $\mathbf{x}^g$ denotes the best solution in the population at the $g$-th generation.
\end{theorem}

\begin{IEEEproof}
Regarding the evolutionary procedure as a finite-dimension Markov chain \cite{Fogel1994Asymptotic}, let
\begin{equation}
\begin{aligned}
s_1:&\ |f(\mathbf{x}^g)-f(\mathbf{x}^\ast)|<\epsilon\\
s_2:&\ |f(\mathbf{x}^g)-f(\mathbf{x}^\ast)|\geq\epsilon
\end{aligned}
\ ,
\end{equation}
then the one-step transition probabilities $P(s_1|s_1)=1$ and $P(s_2|s_1)=0$, since the elite strategy always retains the best solution. Now we focus on the transition probabilities $P(s_1|s_2)$ and $P(s_2|s_2)$. Since $f$ is a continuous function, there must exist a $\varepsilon>0$ such that $|f(\mathbf{x})-f(\mathbf{x}^\ast)|<\epsilon$ when $\|\mathbf{x}-\mathbf{x}^\ast\|_\infty<\varepsilon$, where $\|\mathbf{x}-\mathbf{x}^\ast\|_\infty$ denotes the largest difference between $\mathbf{x}$ and $\mathbf{x}^\ast$ over all the $D$ dimensions. According to equation (\ref{equ:AutoV1}),
\begin{equation}
\begin{aligned}
h(x_{1d},\dots,x_{td})=&\ (1-\sum_{i=2}^{t}r_i)x_{1d}+\sum_{i=2}^{t}r_ix_{id}\\
=&\ x_{1d}+\sum_{i=2}^{t}r_i(x_{id}-x_{1d})\\
=&\ x_{1d}+\sum_{i=2}^{t}r_i^\prime\ ,
\end{aligned}
\end{equation}
where
\begin{equation}
r^\prime_{i+1}\sim\mathcal{N}(\mu_i,\sigma_i^2(x_{(i+1)d}-x_{1d})^2)\ .
\end{equation}
Therefore,
\begin{equation}
\begin{aligned}
&P\left(\|h(\mathbf{x}_1,\dots,\mathbf{x}_t)-\mathbf{x}^\ast\|_\infty<\varepsilon\right)\\
=&\ \prod_{d=1}^{D}P\left(|h(x_{1d},\dots,x_{td})-x_d^\ast|<\varepsilon\right)\\
=&\ \prod_{d=1}^{D}P\left(\left|x_{1d}+\sum_{i=2}^{t}r_{id}^\prime-x_d^\ast\right|<\varepsilon\right)\\
=&\ \prod_{d=1}^{D}P\left(x_d^\ast-x_{1d}-\varepsilon<\sum_{i=2}^{t}r_{id}^\prime<x_d^\ast-x_{1d}+\varepsilon\right)\ .
\end{aligned}
\end{equation}
Since all the $r_{id}^\prime$ are normally distributed,
\begin{equation}
\begin{aligned}
&P\left(\frac{x_d^\ast-x_{1d}-\varepsilon}{t-1}<r_{id}^\prime<\frac{x_d^\ast-x_{1d}+\varepsilon}{t-1}\right)\\
=&\ \int_{\frac{x_d^\ast-x_{1d}-\varepsilon}{t-1}}^{\frac{x_d^\ast-x_{1d}+\varepsilon}{t-1}}\frac{e^{-\frac{c^2}{2\sigma_i^2(x_{(i+1)d}-x_{1d})^2}}}{\sigma_i(x_{(i+1)d}-x_{1d})\sqrt{2\pi}}dc>0\ ,
\end{aligned}
\end{equation}
and thus
\begin{equation}
P\left(\|h(\mathbf{x}_1,\dots,\mathbf{x}_t)-\mathbf{x}^\ast\|_\infty<\varepsilon\right)>0\ ,
\end{equation}
which further indicates that $P(s_1|s_2)>0$. Since $P(s_1|s_2)+P(s_2|s_2)=1$, we have $P(s_2|s_2)<1$. Therefore,
\begin{equation}
\lim_{g\rightarrow\infty}\left[P(s_2|s_2)\right]^g=0\ ,
\end{equation}
which means that the probability of $s_1: |f(\mathbf{x}^g)-f(\mathbf{x}^\ast)|<\epsilon$ when $g\rightarrow\infty$ is 1.
\end{IEEEproof}

It is worth noting that some existing operators include multiple parameter sets to be selected with given probabilities (e.g., $\beta$ in SBX), which provide complex search behaviors and can better balance between exploitation and exploration. Hence, the ensemble of multiple parameter sets is also adopted in AutoV. Moreover, the first weight $r_1$ can be omitted since it can be directly set to $1-r_2-r_3-\cdots$. To summarize, an operator in AutoV is determined by the following matrix
\begin{equation}
\left(
\begin{aligned}
\mu_{12},\sigma_{12},\mu_{13},\sigma_{13},\dots,\mu_{1t},\sigma_{1t},p_1\\
\mu_{22},\sigma_{22},\mu_{23},\sigma_{23},\dots,\mu_{2t},\sigma_{2t},p_2\\
\dots\ \ \dots\ \ \ \ \ \ \ \ \ \ \ \ \ \\
\mu_{k2},\sigma_{k2},\mu_{k3},\sigma_{k3},\dots,\mu_{kt},\sigma_{kt},p_k
\end{aligned}
\right)\ ,
\label{equ:AutoV2}
\end{equation}
where $\mu_{ji}$ and $\sigma_{ji}$ denote the mean and variance of weight $r_i$ in the $j$-th set, respectively, and $p_j$ denotes the probability of selecting the $j$-th set. When generating a decision variable of an offspring solution, the roulette-wheel selection is first used to select a parameter set (i.e., one row of (\ref{equ:AutoV2})) according to their probabilities. Then, the decision variable is generated by sampling the given normal distribution.

In this way, the search of high-performance operators can be formulated as a continuous optimization problem, whose decision vector contains all the elements in (\ref{equ:AutoV2}) and the objective is the performance of the corresponding operator. Generally, it is easy to measure the fitness by investigating the performance of a metaheuristic equipped with the operator, but it is difficult to optimize the decision vector since AutoV does not require any prior knowledge, i.e., none of the existing optimizers are used to evolve a population for solving the problem. To address this issue, the proposed AutoV suggests a novel evolutionary procedure, in which the population can be evolved by itself.

\subsection{Procedure of the Principled Approach}

\begin{figure*}[!t]
  \centering
  \includegraphics[width=1\linewidth]{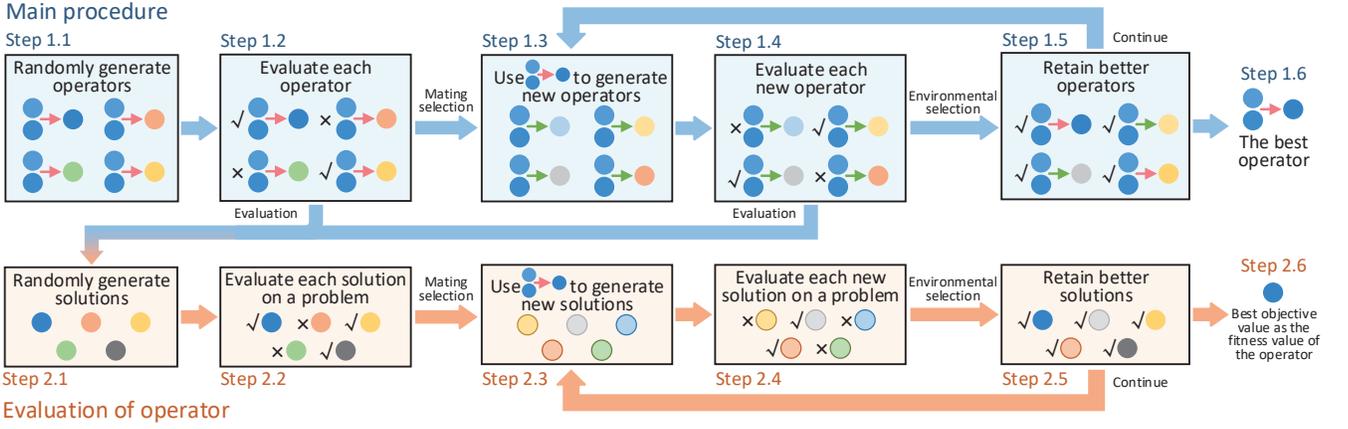}
  \caption{Procedure of AutoV.}
  \label{fig:framework}
\end{figure*}

The procedure of the proposed AutoV is detailed in Fig.~\ref{fig:framework} and Algorithm~\ref{alg:main}. Given a benchmark problem for performance measurement, AutoV first randomly initializes a population $P$ and evaluates each solution in $P$, where each solution denotes a parameter matrix defined in (\ref{equ:AutoV2}). At each generation, AutoV selects a number of parents from $P$ via binary tournament selection, then uses the parents to generate an offspring population by using the variation operator parameterized by the best solution in $P$. Afterwards, the offspring population is combined with $P$, and half the solutions with better fitness survive for the next generation. The proposed AutoV repeats the above steps until the termination criterion is fulfilled, and returns the solution with the best fitness as the found variation operator.

\begin{algorithm}[t]
\caption{Procedure of AutoV}
\label{alg:main}
\SetKwComment{Comment}{//}{}
\KwIn{$f$ (a benchmark problem)}
\KwOut{$\mathbf{p}$ (the best solution)}
$P\leftarrow$ Randomly initialize a population, where each solution is a parameter matrix defined in (\ref{equ:AutoV2})\;
\For{each $\mathbf{x}\in P$}
{
    Evaluate the objective value of $\mathbf{x}$ by $Evaluation(\mathbf{x},f)$\;
}
\While{termination criterion is not fulfilled}
{
    $P^\prime\leftarrow$ Select parents from $P$ via binary tournament selection\;
    $\mathbf{p}\leftarrow$ The best solution in $P$\;
    $O\leftarrow$ Use the operator $\mathbf{p}$ to generate offspring based on parents $P^\prime$\;
    \For{each $\mathbf{o}\in O$}
    {
        Evaluate the objective value of $\mathbf{o}$ by $Evaluation(\mathbf{o},f)$;
    }
    $P\leftarrow P\cup O$\;
    $P\leftarrow$ Half the solutions in $P$ with better fitness\;
}
$\mathbf{p}\leftarrow$ The best solution in $P$\;
\textbf{return} $\mathbf{p}$\;
\end{algorithm}

Since the goal of AutoV is to create high-performance variation operators, it uses the best operator found so far to evolve the population for finding a better operator, and the better operator can then evolve the population for finding a much better operator, where none of the existing metaheuristics are used. For the fitness evaluation of each candidate operator, a simple metaheuristic is established by adopting the candidate operator. As presented in Algorithm~\ref{alg:main2}, the best objective value found by the established metaheuristic on the given benchmark problem is regarded as the fitness of the candidate operator. Besides, we find that a candidate operator may be over-optimized for the given benchmark problem, which means that it considers a perfect solution for the given benchmark problem accidentally. Still, it cannot evolve the population gradually. To solve this issue, AutoV executes the established metaheuristic for multiple runs and uses the median value of the found best objective values as the fitness.

\begin{algorithm}[t]
\caption{$Evaluation(\mathbf{p},f)$}
\label{alg:main2}
\SetKwComment{Comment}{//}{}
\KwIn{$\mathbf{p}$ (parameter matrix of an operator), $f$ (a benchmark problem)}
\KwOut{$fit$ (fitness of $\mathbf{p}$)}
$fit\leftarrow\emptyset$\;
\For{$run=1$ to $maxRun$}
{
    $P\leftarrow$ Randomly initialize a population, where each solution is a decision vector for $f$\;
    \For{each $\mathbf{x}\in P$}
    {
        Evaluate the objective of $\mathbf{x}$ by $f(\mathbf{x})$\;
    }
    \While{termination criterion is not fulfilled}
    {
        $P^\prime\leftarrow$ Select parents from $P$ via binary tournament selection\;
        $O\leftarrow$ Use the operator $\mathbf{p}$ to generate offspring based on parents $P^\prime$\;
        \For{each $\mathbf{o}\in O$}
        {
            Evaluate the objective of $\mathbf{o}$ by $f(\mathbf{o})$;
        }
        $P\leftarrow P\cup O$\;
        $P\leftarrow$ Half the solutions in $P$ with better fitness\;
    }
    $\mathbf{x}\leftarrow$ The best solution in $P$\;
    $fit\leftarrow fit\cup\{f(\mathbf{x})\}$\;
}
$fit\leftarrow$ Median value of $fit$\;
\textbf{return} $fit$\;
\end{algorithm}

In the experiments, we consider the following five different sets of parents as the input of (\ref{equ:AutoV1}) for generating one offspring solution:
\begin{equation}
\left\{
\begin{aligned}
h_1=&\ h(x_{1d},x_{2d})\\
h_2=&\ h(x_{1d},l_d,u_d)\\
h_3=&\ h(x_{1d},x_{2d},l_d,u_d)\\
h_4=&\ h(x_{1d},x_{2d},x_{3d})\\
h_5=&\ h(x_{1d},x_{2d},x_{3d},l_d,u_d)\\
\end{aligned}
\right.,
\label{equ:operatorfunction}
\end{equation}
where $x_{1d},x_{2d},x_{3d}$ denote the decision variables of three parents, $l_d$ denotes the lower bound, and $u_d$ denotes the upper bound on the $d$-th dimension.

\section{Experimental Studies}

\begin{table}[!t]
\renewcommand{\arraystretch}{1.2}
\footnotesize
\centering
\caption{Parameter Settings of the Compared Metaheuristics, Where $D$ is the Number of Decision Variables, $\mathbf{u}$ is the Upper Bound, $\mathbf{l}$ is the Lower Bound, and $n=100$ is the Population Size.}
\label{tab:setting}
\setlength{\tabcolsep}{0.8mm}{
\begin{tabular}{c|l}
\hline
Metaheuristic&\makebox[5cm][c]{\ \ \ \ \ \ \ Parameter setting}\\
\hline
GA \cite{Holland1992Adaptation}&crossover probability: 1,\\
(based on SBX \cite{Operator-SBX} and&mutation probability: $1/D$,\\
polynomial mutation \cite{Operator-polynomial-mutation})&distribution index: 20\\
\hline
PSO \cite{Operator-PSO}&inertia weight: 0.4\\
\hline
DE \cite{Operator-DE}&\MR2{$CR=0.9$, $F=0.5$}\\
(DE/rand/1/bin)&\\
\hline
&initial $\sigma=0.6(\mathbf{u}-\mathbf{l})$,\\
\MR2{CMA-ES \cite{algorithm-CMAES}}&$w_i=\log(\mu+0.5)-\log(i)$, $i\in[1,\mu]$,\\
&$\mu=0.5n$, $c_\sigma=0.15$,\\
&$d_\sigma=1.15$, $c_c=0.105$\\
\hline
FEP \cite{Operator-FEP}&initial $\eta=3$\\
\hline
CSO \cite{algorithm-CSO}&\multirow{3}{*}{social factor: 0.1}\\
(competitive swarm&\\
optimizer)&\\
\hline
SHADE \cite{Algorithm-SHADE}&\MR2{\ --}\\
(parameter adaptative DE)&\\
\hline
IMODE \cite{Algorithm-IMODE}&Minimum population size: 4\\
(winner of CEC'2020)&Ratio of archive size: 2.6\\
\hline
\end{tabular}
}
\end{table}

To verify the effectiveness of the proposed AutoV, the performance of the operators found by AutoV is first studied. Then, the best operator is compared with eight metaheuristics, where GA \cite{Holland1992Adaptation}, PSO \cite{Operator-PSO}, DE \cite{Operator-DE}, CMA-ES \cite{algorithm-CMAES}, and FEP \cite{Operator-FEP} are classical metaheuristics, CSO \cite{algorithm-CSO} is a competitive swarm optimizer for large-scale optimization, and SHADE \cite{Algorithm-SHADE} and IMODE \cite{Algorithm-IMODE} are hybrid metaheuristics based on multiple operators with parameter adaptation. Based on the settings suggested in the original literature of the compared metaheuristics, we finely tune their parameters for a relatively good performance, where the detailed parameter settings are listed in Table~\ref{tab:setting}.

\subsection{Comparison Between the Operators Found by AutoV}

For each of the five functions given in (\ref{equ:operatorfunction}), the proposed AutoV optimizes it with a population size of 100 for 1000 generations. As for the fitness evaluation of each candidate operator, the population size is set to 100, the number of generations is set to 100, the number of parameter sets $k$ is set to 10, and the number of runs $maxRun$ is set to 9. Besides, the Rastrigin's Function \cite{Su2020Non} is adopted as the benchmark problem for performance measurement.

To compare the performance of the found operators with different functions $h_1$--$h_5$, they are embedded in the simple metaheuristic presented in Algorithm~\ref{alg:main2} and tested on eight benchmark problems with 30 decision variables. These benchmark problems have a variety of unimodal, multimodal, or flat landscapes, whose definitions can be found in \cite{Su2020Non}. For all the metaheuristics, the population size is set to 100 and the number of generations is set to 100. Table~\ref{tab:exp0} lists the minimum objective values found by the five metaheuristics averaged over 30 runs, where the compared operators exhibit similar performance and the operator $h_3$ has slightly better overall performance than the others. It is worth noting that the operator $h_5$ does not obtain the best overall performance, though the function $h_5$ has more parents and is expected to perform better. This is because the function $h_5$ contains more parameters to be optimized, which hinders AutoV from finding high-performance operators.

\begin{table}[!t]
\renewcommand{\arraystretch}{1.2}
\footnotesize
\centering
\caption{Minimum Objective Values Obtained by the Operators with Different Functions $h_1$--$h_5$ Found by AutoV on Eight Benchmark Problems. Best Results are Highlighted.}
\label{tab:exp0}
\setlength{\tabcolsep}{1.2mm}{
\begin{tabular}{cccccc}
\toprule
Problem&$h_1$&$h_2$&$h_3$&$h_4$&$h_5$\\
\midrule
Schwefel's&\MR2{\hl{.07e-5}}&\MR2{5.27e+0}&\MR2{2.07e-2}&\MR2{8.72e-3}&\MR2{1.02e-1}\\
Function 2.22\\
\hline
Schwefel's&\MR2{1.65e+1}&\MR2{1.19e+1}&\MR2{\hl{1.66e+0}}&\MR2{8.75e+0}&\MR2{1.67e+0}\\
Function 2.21\\
\hline
Quartic&\MR2{2.01e-2}&\MR2{1.32e-1}&\MR2{1.12e-2}&\MR2{\hl{9.02e-3}}&\MR2{1.16e-2}\\
Function\\
\hline
Generalized\\
Griewank&\hl{1.99e-3}&2.93e+0&2.81e-2&4.46e-1&2.66e-1\\
Function\\
\hline
Generalized\\
Schwefel's&-9.17e+3&-1.04e+4&\hl{-1.16e+4}&-5.70e+3&-5.28e+3\\
Function 2.26\\
\hline
Ackley's&\MR2{\hl{5.47e-4}}&\MR2{4.77e+0}&\MR2{1.79e-2}&\MR2{8.94e-1}&\MR2{2.05e-1}\\
Function\\
\hline
Rosenbrock's&\MR2{1.20e+4}&\MR2{1.17e+5}&\MR2{9.09e+3}&\MR2{8.80e+3}&\MR2{\hl{3.82e+3}}\\
Function\\
\hline
Rastrigin's&\MR2{3.76e+1}&\MR2{3.96e+1}&\MR2{\hl{3.88e+0}}&\MR2{9.99e+0}&\MR2{7.86e+0}\\
Function\\
\bottomrule
\end{tabular}
}
\end{table}

\begin{figure}[!t]
  \centering
  \subfloat{\includegraphics[width=1\linewidth]{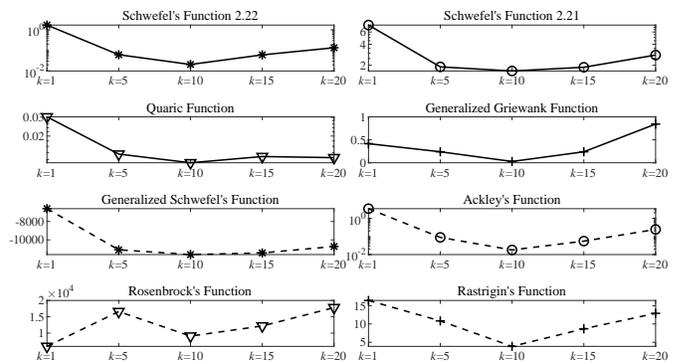}}
  \caption{Minimum objective values obtained by the operator with function $h_3$ and different numbers of parameter sets $k$ found by AutoV.}
  \label{fig:exp0}
\end{figure}

\begin{table}[!t]
\renewcommand{\arraystretch}{0.9}
\scriptsize
\centering
\caption{Minimum Objective Values Obtained by the Operator with Function $h_3$ and Different Benchmark Problems for Fitness Evaluation. Best Results are Highlighted.}
\label{tab:exp02}
\setlength{\tabcolsep}{0.8mm}{
\begin{tabular}{ccccc}
\toprule
\multirow{3}{*}{Problem}&Optimized on&Optimized on&Optimized on&Optimized on\\
&Schwefel's&Schwefel's&Ackley's&Rastrigin's\\
&Function 2.22&Function 2.21&Function&Function\\
\midrule
Schwefel's&\multirow{2}{*}{\hl{2.46e-6}}&\multirow{2}{*}{7.00e-2}&\multirow{2}{*}{4.91e-6}&\multirow{2}{*}{2.07e-2}\\
Function 2.22\\
\hline
Schwefel's&\multirow{2}{*}{1.46e+1}&\multirow{2}{*}{\hl{1.36e+0}}&\multirow{2}{*}{1.21e+1}&\multirow{2}{*}{1.66e+0}\\
Function 2.21\\
\hline
Quartic&\multirow{2}{*}{1.29e-2}&\multirow{2}{*}{1.23e-2}&\multirow{2}{*}{1.44e-2}&\multirow{2}{*}{\hl{1.12e-2}}\\
Function\\
\hline
Generalized&\multirow{3}{*}{6.89e-3}&\multirow{3}{*}{7.02e-2}&\multirow{3}{*}{\hl{1.48e-3}}&\multirow{3}{*}{2.81e-2}\\
Griewank\\
Function\\
\hline
Generalized&\multirow{3}{*}{-9.35e+3}&\multirow{3}{*}{-9.76e+3}&\multirow{3}{*}{-9.55e+3}&\multirow{3}{*}{\hl{-1.16e+4}}\\
Schwefel's\\
Function 2.26\\
\hline
Ackley's&\multirow{2}{*}{1.00e-4}&\multirow{2}{*}{5.77e-2}&\multirow{2}{*}{\hl{5.71e-5}}&\multirow{2}{*}{1.79e-2}\\
Function\\
\hline
Rosenbrock's&\multirow{2}{*}{6.03e+3}&\multirow{2}{*}{\hl{3.89e+3}}&\multirow{2}{*}{7.64e+3}&\multirow{2}{*}{9.09e+3}\\
Function\\
\hline
Rastrigin's&\multirow{2}{*}{2.36e+1}&\multirow{2}{*}{1.59e+1}&\multirow{2}{*}{2.91e+1}&\multirow{2}{*}{\hl{3.88e+0}}\\
Function\\
\bottomrule
\end{tabular}
}
\end{table}

\begin{table*}[!t]
\renewcommand{\arraystretch}{0.7}
\centering
\caption{Minimum Objective Values Obtained by Nine Metaheuristics on 13 Small-Scale Benchmark Problems with 30 Decision Variables. Best Results are Highlighted.}
\label{tab:exp1}
\resizebox{\textwidth}{!}{
\begin{tabular}{cccccccccc}
\toprule
Small-scale&\MR2{GA \cite{Holland1992Adaptation}}&\MR2{PSO \cite{Operator-PSO}}&\MR2{DE \cite{Operator-DE}}&\MR2{CMA-ES \cite{algorithm-CMAES}}&\MR2{FEP \cite{Operator-FEP}}&\MR2{CSO \cite{algorithm-CSO}}&\MR2{SHADE \cite{Algorithm-SHADE}}&\MR2{IMODE \cite{Algorithm-IMODE}}&\MR2{AutoV}\\
problems \cite{Operator-FEP}&\\
\midrule
\multirow{2}{*}{$f_1$}&1.4973e+1 $-$&3.5453e+3 $-$&5.0032e+3 $-$&1.2391e+2 $-$&9.2809e+3 $-$&1.2988e+3 $-$&2.2958e+1 $-$&2.9105e+0 $-$&\hl{8.3702e-1}\\
&(2.0053e+0)&(1.2242e+3)&(5.7205e+2)&(2.6004e+1)&(2.7503e+3)&(5.0772e+2)&(7.4054e+0)&(9.0907e-1)&\hl{(5.333e-1)}\\[1mm]

\multirow{2}{*}{$f_2$}&8.3310e-1 $-$&3.2170e+1 $-$&5.1595e+1 $-$&8.6103e+0 $-$&5.1291e+1 $-$&1.5504e+1 $-$&6.2312e+0 $-$&3.2305e-1 $-$&\hl{1.2081e-1}\\
&(1.7114e-1)&(6.6473e+0)&(1.5480e+1)&(3.3770e+0)&(1.0920e+1)&(3.1662e+0)&(1.1158e+0)&(1.0412e-1)&\hl{(2.556e-2)}\\[1mm]

\multirow{2}{*}{$f_3$}&1.0621e+4 $-$&7.8076e+3 $-$&6.4156e+3 $-$&5.0816e+3 $-$&2.1959e+4 $-$&1.7542e+3 $\approx$&1.6461e+3 $\approx$&\hl{6.6967e+2 $+$}&2.0131e+3\\
&(2.6732e+3)&(2.9985e+3)&(1.1614e+3)&(1.2254e+3)&(3.5707e+3)&(5.5302e+2)&(5.3479e+2)&\hl{(1.3529e+2)}&(8.307e+2)\\[1mm]

\multirow{2}{*}{$f_4$}&2.3222e+1 $-$&2.4005e+1 $-$&3.4503e+1 $-$&8.7765e+0 $-$&5.3464e+1 $-$&1.1521e+1 $-$&9.1224e+0 $-$&8.8346e+0 $-$&\hl{6.4179e+0}\\
&(3.6870e+0)&(3.0366e+0)&(4.1625e+0)&(1.5806e+0)&(9.3989e+0)&(2.0034e+0)&(1.4948e+0)&(2.0574e+0)&\hl{(1.953e+0)}\\[1mm]

\multirow{2}{*}{$f_5$}&1.2336e+3 $-$&5.5668e+5 $-$&1.6919e+6 $-$&5.1996e+3 $-$&5.8033e+6 $-$&1.5258e+5 $-$&7.2347e+2 $-$&\hl{1.2120e+2 $\approx$}&2.0580e+2\\
&(6.8051e+2)&(2.7471e+5)&(1.0907e+6)&(2.8736e+3)&(2.9157e+6)&(7.7532e+4)&(3.3376e+2)&\hl{(6.9918e+1)}&(1.981e+2)\\[1mm]

\multirow{2}{*}{$f_6$}&1.9375e+1 $-$&3.4820e+3 $-$&4.5314e+3 $-$&1.3163e+2 $-$&8.8923e+3 $-$&1.0266e+3 $-$&3.2250e+1 $-$&3.6250e+0 $-$&\hl{0.0000e+0}\\
&(3.9978e+0)&(1.1301e+3)&(8.0852e+2)&(3.2824e+1)&(3.1734e+3)&(4.3859e+2)&(7.7965e+0)&(2.1998e+0)&\hl{(0.000e+0)}\\[1mm]

\multirow{2}{*}{$f_7$}&9.1447e-2 $-$&8.2203e-1 $-$&1.1222e+0 $-$&8.2239e-2 $-$&5.2040e+1 $-$&5.4577e-2 $-$&5.8113e-2 $-$&8.4790e-2 $-$&\hl{2.0943e-2}\\
&(3.8031e-2)&(2.2065e-1)&(3.6673e-1)&(2.6534e-2)&(2.7550e+1)&(2.1989e-2)&(1.4618e-2)&(2.7932e-2)&\hl{(8.192e-3)}\\[1mm]

\multirow{2}{*}{$f_8$}&-1.4025e+4 $\approx$&-8.2308e+3 $-$&-1.1684e+4 $-$&-1.2799e+4 $-$&-1.0531e+4 $-$&-1.2285e+4 $-$&-1.4023e+4 $\approx$&\hl{-1.4575e+4 $+$}&-1.3852e+4\\
&(5.0347e+2)&(1.2979e+3)&(5.6921e+2)&(2.8077e+2)&(8.2223e+2)&(4.2013e+2)&(8.5233e+1)&\hl{(2.0746e+2)}&(6.717e+2)\\[1mm]

\multirow{2}{*}{$f_9$}&\hl{1.1504e+1 $+$}&1.5864e+2 $-$&2.7333e+2 $-$&2.2788e+2 $-$&2.6914e+2 $-$&1.0901e+2 $-$&1.6771e+2 $-$&2.3842e+1 $-$&1.4546e+1\\
&\hl{(3.3186e+0)}&(1.7533e+1)&(1.4085e+1)&(2.3562e+1)&(2.7653e+1)&(1.7125e+1)&(1.4089e+1)&(5.6385e+0)&(2.163e+0)\\[1mm]

\multirow{2}{*}{$f_{10}$}&1.7023e+0 $-$&1.2329e+1 $-$&1.3329e+1 $-$&4.0097e+0 $-$&1.4671e+1 $-$&7.6077e+0 $-$&2.8370e+0 $-$&1.2297e+0 $-$&\hl{2.1621e-1}\\
&(3.1651e-1)&(1.3322e+0)&(8.5004e-1)&(5.6144e-1)&(7.3884e-1)&(1.1400e+0)&(2.1827e-1)&(3.8807e-1)&\hl{(5.119e-2)}\\[1mm]

\multirow{2}{*}{$f_{11}$}&1.1563e+0 $-$&2.8566e+1 $-$&4.6051e+1 $-$&2.1092e+0 $-$&1.1710e+2 $-$&1.0340e+1 $-$&1.2644e+0 $-$&1.0158e+0 $-$&\hl{6.7928e-1}\\
&(4.0584e-2)&(3.2736e+0)&(1.3901e+1)&(2.8414e-1)&(3.2432e+1)&(3.2046e+0)&(7.1606e-2)&(3.6145e-2)&\hl{(1.999e-1)}\\[1mm]

\multirow{2}{*}{$f_{12}$}&2.8786e+1 $-$&1.0624e+3 $-$&7.1756e+4 $-$&4.0254e+1 $-$&4.9839e+6 $-$&1.3627e+2 $-$&3.1790e+1 $-$&7.2029e+1 $-$&\hl{1.5847e+0}\\
&(1.5748e+1)&(1.3982e+3)&(6.5424e+4)&(5.5667e+0)&(5.9492e+6)&(3.5766e+1)&(9.1497e+0)&(2.4378e+1)&\hl{(1.976e+0)}\\[1mm]

\multirow{2}{*}{$f_{13}$}&7.1895e+0 $-$&4.3602e+5 $-$&2.1898e+6 $-$&1.0939e+1 $-$&1.2093e+7 $-$&2.1808e+3 $-$&4.3094e+0 $-$&4.3502e+0 $-$&\hl{1.0075e-1}\\
&(7.2480e+0)&(5.1304e+5)&(1.7255e+6)&(2.7540e+0)&(1.0145e+7)&(5.2706e+3)&(1.5238e+0)&(2.3093e+0)&\hl{(4.216e-2)}\\[1mm]
\hline
\multicolumn{1}{c}{$+/-/\approx$}&1/11/1&0/13/0&0/13/0&0/13/0&0/13/0&0/12/1&0/11/2&2/10/1&\\
\bottomrule
\end{tabular}
}
\begin{tablenotes}
\footnotesize
\item["\dag"] '$+$', '$-$' and '$\approx$' indicate that the result is significantly better, significantly worse, and statistically similar to that obtained by AutoV.
\end{tablenotes}
\end{table*}

To study the influence of the number of parameter sets $k$, Fig.~\ref{fig:exp0} plots the performance of the metaheuristic with operator $h_3$ and $k=1,5,10,15,20$ on the eight benchmark problems, where $k=10$ leads to better overall performance than the other settings. On the one hand, a small value of $k$ provides a few different search behaviors, and thus leads to a low performance limit. On the other hand, although a large value of $k$ provides many different search behaviors, it leads to a large number of parameters that are difficult to be optimized. As a consequence, $k=10$ is a proper setting for finding high-performance operators.

\begin{table*}[!t]
\renewcommand{\arraystretch}{0.7}
\centering
\caption{Minimum Objective Values Obtained by Nine Metaheuristics on 13 Rotated Small-Scale Benchmark Problems with 30 Decision Variables. Best Results are Highlighted.}
\label{tab:exp2}
\resizebox{\textwidth}{!}{
\begin{tabular}{cccccccccc}
\toprule
Rotated&\\
small-scale&GA \cite{Holland1992Adaptation}&PSO \cite{Operator-PSO}&DE \cite{Operator-DE}&CMA-ES \cite{algorithm-CMAES}&FEP \cite{Operator-FEP}&CSO \cite{algorithm-CSO}&SHADE \cite{Algorithm-SHADE}&IMODE \cite{Algorithm-IMODE}&AutoV\\
problems \cite{Operator-FEP}&\\
\midrule
\multirow{2}{*}{$f_1$}&1.7644e+1 $-$&3.1210e+3 $-$&4.3448e+3 $-$&1.2422e+2 $-$&9.2516e+3 $-$&1.2894e+3 $-$&2.9700e+1 $-$&2.0650e+0 $-$&\hl{7.1625e-1}\\
&(5.3457e+0)&(8.2974e+2)&(8.3965e+2)&(2.9570e+1)&(3.9378e+3)&(6.7999e+2)&(1.0900e+1)&(1.0838e+0)&\hl{(5.403e-1)}\\[1mm]

\multirow{2}{*}{$f_2$}&1.9315e+1 $-$&3.7244e+1 $-$&4.9671e+1 $-$&1.3854e+1 $-$&8.1606e+1 $-$&1.4780e+1 $-$&1.8746e+1 $-$&1.5749e+0 $\approx$&\hl{1.5246e+0}\\
&(1.0234e+1)&(1.8184e+1)&(8.5749e+0)&(6.8676e+0)&(9.7066e+0)&(3.0538e+0)&(2.6580e+0)&(4.2566e-1)&\hl{(5.190e-1)}\\[1mm]

\multirow{2}{*}{$f_3$}&9.9581e+3 $-$&5.3590e+3 $-$&5.3369e+3 $-$&1.3929e+4 $-$&2.2100e+4 $-$&1.5239e+3 $\approx$&1.7530e+3 $\approx$&\hl{8.5084e+2 $+$}&2.1193e+3\\
&(2.6366e+3)&(1.4792e+3)&(1.3071e+3)&(3.9351e+3)&(5.0673e+3)&(9.7521e+2)&(4.7904e+2)&\hl{(2.5578e+2)}&(7.960e+2)\\[1mm]

\multirow{2}{*}{$f_4$}&1.4962e+1 $-$&2.4756e+1 $-$&3.6628e+1 $-$&7.8044e+0 $-$&4.6193e+1 $-$&1.2863e+1 $-$&8.1105e+0 $-$&5.5831e+0 $-$&\hl{3.2227e+0}\\
&(5.9860e+0)&(4.8244e+0)&(4.7317e+0)&(9.4482e-1)&(2.4146e+0)&(1.8236e+0)&(8.4983e-1)&(1.2362e+0)&\hl{(1.742e+0)}\\[1mm]

\multirow{2}{*}{$f_5$}&2.8300e+4 $-$&8.9721e+5 $-$&1.7625e+6 $-$&9.3046e+3 $-$&5.2830e+6 $-$&1.1350e+5 $-$&2.8513e+3 $-$&2.9329e+2 $\approx$&\hl{2.0175e+2}\\
&(4.8151e+4)&(3.4766e+5)&(7.4240e+5)&(6.0657e+3)&(2.3632e+6)&(6.7006e+4)&(3.4473e+3)&(1.6130e+2)&\hl{(1.540e+2)}\\[1mm]

\multirow{2}{*}{$f_6$}&2.3375e+1 $-$&3.6026e+3 $-$&5.0828e+3 $-$&1.1925e+2 $-$&8.6534e+3 $-$&1.2864e+3 $-$&3.2250e+1 $-$&1.3750e+1 $-$&\hl{2.5000e+0}\\
&(3.7009e+0)&(6.8520e+2)&(1.1280e+3)&(2.9011e+1)&(2.1819e+3)&(5.8652e+2)&(6.5629e+0)&(7.1464e+0)&\hl{(2.204e+0)}\\[1mm]

\multirow{2}{*}{$f_7$}&8.1742e-2 $-$&8.6616e-1 $-$&1.1051e+0 $-$&7.1672e-2 $-$&4.8419e+1 $-$&1.1136e-1 $-$&6.3319e-2 $-$&4.9230e-2 $-$&\hl{1.7558e-2}\\
&(2.9263e-2)&(4.3408e-1)&(6.6114e-1)&(3.4086e-2)&(2.8474e+1)&(5.2390e-2)&(1.6838e-2)&(2.2971e-2)&\hl{(5.800e-3)}\\[1mm]

\multirow{2}{*}{$f_8$}&\hl{-7.6960e+3 $\approx$}&-6.4301e+3 $-$&-5.2233e+3 $-$&-7.6329e+3 $\approx$&-7.4567e+3 $\approx$&-7.5311e+3 $\approx$&-5.5305e+3 $-$&-7.1206e+3 $\approx$&-7.3751e+3\\
&\hl{(4.2092e+2)}&(7.3448e+2)&(4.2305e+2)&(5.6710e+2)&(6.5244e+2)&(2.9851e+2)&(2.9295e+2)&(2.0444e+2)&(2.951e+2)\\[1mm]

\multirow{2}{*}{$f_9$}&7.0727e+1 $-$&1.4357e+2 $-$&2.6278e+2 $-$&2.2499e+2 $-$&3.0943e+2 $-$&1.0129e+2 $-$&2.1434e+2 $-$&8.0586e+1 $-$&\hl{4.1289e+1}\\
&(1.4655e+1)&(1.9674e+1)&(1.3668e+1)&(2.5540e+1)&(1.3695e+1)&(1.1932e+1)&(1.3079e+1)&(1.7108e+1)&\hl{(1.364e+1)}\\[1mm]

\multirow{2}{*}{$f_{10}$}&2.9250e+0 $-$&1.1749e+1 $-$&1.3413e+1 $-$&4.3282e+0 $-$&1.5657e+1 $-$&7.3892e+0 $-$&3.0076e+0 $-$&2.3761e+0 $-$&\hl{9.6926e-1}\\
&(1.0910e-1)&(5.8613e-1)&(8.0644e-1)&(4.3019e-1)&(1.5809e+0)&(9.1288e-1)&(3.2128e-1)&(5.8902e-1)&\hl{(6.627e-1)}\\[1mm]

\multirow{2}{*}{$f_{11}$}&1.1551e+0 $-$&2.9789e+1 $-$&4.5249e+1 $-$&1.9904e+0 $-$&1.1321e+2 $-$&1.2782e+1 $-$&1.1978e+0 $-$&9.4447e-1 $-$&\hl{7.6236e-1}\\
&(2.4035e-2)&(8.3839e+0)&(8.3656e+0)&(2.2665e-1)&(2.2152e+1)&(4.1098e+0)&(7.2707e-2)&(7.3867e-2)&\hl{(1.343e-1)}\\[1mm]

\multirow{2}{*}{$f_{12}$}&3.5756e+1 $-$&1.5007e+3 $-$&1.9844e+5 $-$&3.7031e+1 $-$&1.2756e+6 $-$&1.3627e+2 $-$&3.4071e+1 $-$&7.2069e+1 $-$&\hl{1.4346e+0}\\
&(1.7714e+1)&(1.6007e+3)&(1.8928e+5)&(6.1593e+0)&(1.1497e+6)&(6.7341e+1)&(5.6141e+0)&(1.6143e+1)&\hl{(1.133e+0)}\\[1mm]

\multirow{2}{*}{$f_{13}$}&4.1967e+0 $-$&6.7210e+5 $-$&3.8989e+6 $-$&1.0194e+1 $-$&9.2196e+6 $-$&7.5595e+3 $-$&5.1567e+0 $-$&6.4904e+0 $-$&\hl{1.4557e-1}\\
&(3.5806e+0)&(4.4019e+5)&(2.7548e+6)&(3.2998e+0)&(7.2385e+6)&(2.0845e+4)&(1.6990e+0)&(1.5676e+0)&\hl{(6.230e-2)}\\[1mm]
\hline
\multicolumn{1}{c}{$+/-/\approx$}&0/12/1&0/13/0&0/13/0&0/12/1&0/12/1&0/11/2&0/12/1&1/9/3&\\
\bottomrule
\end{tabular}
}
\begin{tablenotes}
\footnotesize
\item["\dag"] '$+$', '$-$' and '$\approx$' indicate that the result is significantly better, significantly worse, and statistically similar to that obtained by AutoV.
\end{tablenotes}
\end{table*}

Furthermore, the influence of the benchmark problem on fitness evaluation is studied. Table~\ref{tab:exp02} lists the performance of the metaheuristic with operator $h_3$ optimized on four benchmark problems, including the Schwefel's Function 2.22 with a unimodal landscape, the Schwefel's Function 2.21 with a flat landscape, and the Ackley's Function and Rastrigin's Function with multimodal landscapes. It can be found that the four metaheuristics obtain the best performance on the benchmark problem for fitness evaluation, while they obtain quite similar performance on the other problems. In the following experiments, the metaheuristic with operator $h_3$ and $k=10$ optimized on the Rastrigin's Function is used as a representative of AutoV to be compared with existing metaheuristics on various problems. Details of the variation operator found by AutoV are presented in the following, where the components with $r_i\sim\mathcal{N}(0,0)$ are ignored for simplicity:
\begin{equation}
\scriptsize
\begin{aligned}
&h(x_1,x_2,l,u)=\\
&\begin{aligned}
\left\{
\begin{aligned}
&(1-r_2)x_1+r_2x_2,&r_2\sim\mathcal{N}(0.4753,0.0103)\\
&&{\rm if}\ p<0.219\\
&(1-r_2)x_1+r_2x_2,&r_2\sim\mathcal{N}(0.0034,0.0006)\\
&&{\rm if}\ 0.219\leq p<0.436\\
&(1-r_2)x_1+r_2x_2,&r_2\sim\mathcal{N}(0.9999,0.0072)\\
&&{\rm if}\ 0.436\leq p<0.644\\
&(1-r_2)x_1+r_2x_2,&r_2\sim\mathcal{N}(0.9988,0.0167)\\
&&{\rm if}\ 0.644\leq p<0.802\\
&(1-r_2)x_1+r_2x_2,&r_2\sim\mathcal{N}(-0.0070,0.0056)\\
&&{\rm if}\ 0.802\leq p<0.960\\
&(1-r_2-r_3-r_4)x_1+r_2x_2+r_3l+r_4u,&r_2\sim\mathcal{N}(0.7293,0.8055)\\
&&r_3\sim\mathcal{N}(0,0.0014)\\
&&r_4\sim\mathcal{N}(0,0.0014)\\
&&{\rm if}\ 0.960\leq p<0.992\\
&(1-r_2-r_3-r_4)x_1+r_2x_2+r_3l+r_4u,&r_2\sim\mathcal{N}(0.0392,0.3185)\\
&&r_3\sim\mathcal{N}(0,0.0351)\\
&&r_4\sim\mathcal{N}(0,0.0351)\\
&&{\rm if}\ 0.992\leq p<0.997\\
&(1-r_2-r_3-r_4)x_1+r_2x_2+r_3l+r_4u,&r_2\sim\mathcal{N}(-0.9998,0.2478)\\
&&r_3\sim\mathcal{N}(0,0.0011)\\
&&r_4\sim\mathcal{N}(0,0.0011)\\
&&{\rm if}\ 0.997\leq p<0.998\\
&(1-r_2-r_3-r_4)x_1+r_2x_2+r_3l+r_4u,&r_2\sim\mathcal{N}(-0.8547,0.1174)\\
&&r_3\sim\mathcal{N}(0,0.0452)\\
&&r_4\sim\mathcal{N}(0,0.0452)\\
&&{\rm if}\ 0.998\leq p<0.999\\
&(1-r_2-r_3-r_4)x_1+r_2x_2+r_3l+r_4u,&r_2\sim\mathcal{N}(-0.5621,0.1059)\\
&&r_3\sim\mathcal{N}(0,0.0003)\\
&&r_4\sim\mathcal{N}(0,0.0003)\\
&&{\rm if}\ p\geq0.999\\
\end{aligned}
\right.,
\end{aligned}
\end{aligned}
\end{equation}
where $p$ is a uniformly distributed random value sampled in $[0,1]$. It can be found from the definition that the operator contains ten functions whose parameters obey different distributions, which provide a variety of search behaviors that may be effective for different problems. It is worth noting that the ten functions are selected with different probabilities, where the operator uses only $x_1$ and $x_2$ in most cases but rarely uses the lower bound $l$ and the upper bound $u$. This is similar to the genetic algorithm, which performs the crossover between two parents with a large probability for global search and performs the mutation on a single solution with a small probability for local search, where the lower and upper bounds are involved in the mutation operator (e.g., polynomial mutation \cite{Operator-polynomial-mutation}) but ignored in the crossover operator (e.g., SBX \cite{Operator-SBX}).

\subsection{Comparison on Small-Scale Benchmark Problems}

\begin{figure}[!t]
  \centering
  \subfloat{\includegraphics[width=1\linewidth]{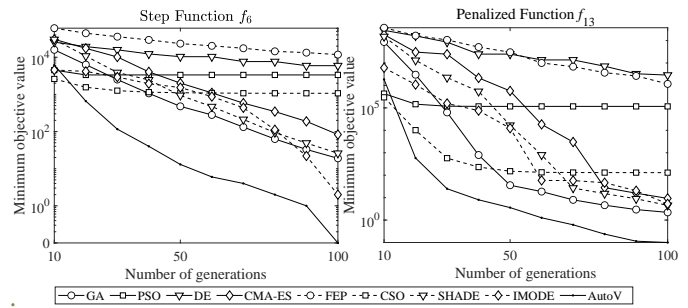}}
  \caption{Convergence trajectories of nine metaheuristics on two problems.}
  \label{fig:exp1}
\end{figure}

Then, the proposed AutoV (i.e., the metaheuristic with operator $h_3$) is compared with eight existing metaheuristics on 13 small-scale benchmark problems with 30 decision variables, where the definitions of these problems can be found in \cite{Operator-FEP}. For all the compared metaheuristics, the population size is set to 100 and the number of function evaluations is set to 10000. Table~\ref{tab:exp1} shows the means and standard deviations of the minimum objective values found by the nine metaheuristics, averaged over 30 runs. It can be found that the proposed AutoV obtains the best overall performance, which gains the best results on 9 out of 13 problems. The Wilcoxon rank sum test \cite{Derrac2011Practical} with a significance level of 0.05 is adopted to perform statistical analysis, where AutoV is significantly better than GA, PSO, DE, CMA-ES, FEP, CSO, SHADE, IMODE on 11, 13, 13, 13, 13, 12, 11, and 10 problems, respectively. Furthermore, Fig.~\ref{fig:exp1} depicts the convergence trajectories of the compared metaheuristics on the Step Function and the Penalized Function, where AutoV converges faster than the other metaheuristics.

\begin{table*}[!t]
\renewcommand{\arraystretch}{0.7}
\centering
\caption{Minimum Objective Values Obtained by Nine Algorithms on the CEC'2013 Large-Scale Benchmark Problems with About 1000 Decision Variables. Best Results are Highlighted.}
\label{tab:exp3}
\resizebox{\textwidth}{!}{
\begin{tabular}{cccccccccc}
\toprule
CEC'2013\\
large-scale&GA \cite{Holland1992Adaptation}&PSO \cite{Operator-PSO}&DE \cite{Operator-DE}&CMA-ES \cite{algorithm-CMAES}&FEP \cite{Operator-FEP}&CSO \cite{algorithm-CSO}&SHADE \cite{Algorithm-SHADE}&IMODE \cite{Algorithm-IMODE}&AutoV\\
problems \cite{li2013benchmark}\\
\midrule
\multirow{2}{*}{$f_1$}&1.1824e+9 $-$&1.7274e+11 $-$&1.1820e+11 $-$&1.3230e+10 $-$&6.2225e+10 $-$&1.6506e+11 $-$&9.2604e+8 $-$&1.5672e+10 $-$&\hl{7.5437e+8}\\
&(1.5088e+8)&(8.2047e+9)&(5.7453e+9)&(5.9476e+8)&(8.8355e+9)&(6.4000e+9)&(9.2525e+7)&(1.8277e+9)&\hl{(3.599e+7)}\\[1mm]

\multirow{2}{*}{$f_2$}&\hl{7.0854e+3 $+$}&4.8900e+4 $-$&4.2656e+4 $-$&1.1326e+4 $+$&8.4194e+4 $-$&4.4704e+4 $-$&1.8598e+4 $-$&2.4830e+4 $-$&1.2736e+4\\
&\hl{(5.5474e+2)}&(1.6515e+3)&(9.8044e+2)&(2.6942e+2)&(5.6859e+2)&(7.5103e+2)&(1.1309e+3)&(6.9359e+2)&(3.996e+2)\\[1mm]

\multirow{2}{*}{$f_3$}&1.9790e+1 $-$&2.1096e+1 $-$&2.1011e+1 $-$&2.1334e+1 $-$&2.1457e+1 $-$&2.0977e+1 $-$&1.7116e+1 $-$&2.0013e+1 $-$&\hl{7.6972e+0}\\
&(1.0792e-1)&(3.6567e-2)&(6.4041e-3)&(2.0615e-2)&(8.6678e-3)&(1.9539e-2)&(2.1804e-1)&(1.6968e-2)&\hl{(2.709e-1)}\\[1mm]

\multirow{2}{*}{$f_4$}&7.0214e+11 $-$&3.8264e+12 $-$&2.8759e+11 $-$&1.8046e+12 $-$&7.4290e+11 $-$&2.0687e+12 $-$&\hl{4.1680e+10 $+$}&3.5605e+11 $-$&1.7434e+11\\
&(3.6361e+11)&(1.0018e+12)&(4.4452e+10)&(2.2407e+11)&(4.1395e+11)&(5.9504e+11)&\hl{(7.7120e+9)}&(3.1439e+10)&(3.356e+10)\\[1mm]

\multirow{2}{*}{$f_5$}&8.0929e+6 $-$&3.3051e+7 $-$&1.6648e+7 $-$&1.1814e+7 $-$&2.0252e+7 $-$&2.7747e+7 $-$&6.6297e+6 $-$&1.0851e+7 $-$&\hl{5.5157e+6}\\
&(1.1105e+6)&(3.3554e+6)&(1.2160e+6)&(3.5863e+5)&(1.9684e+6)&(2.3903e+6)&(8.5226e+5)&(6.3028e+5)&\hl{(7.817e+5)}\\[1mm]

\multirow{2}{*}{$f_6$}&7.6705e+5 $-$&1.0156e+6 $-$&9.6778e+5 $-$&1.0725e+6 $-$&7.7643e+5 $-$&9.9765e+5 $-$&1.0802e+5 $-$&2.7032e+5 $-$&\hl{8.3894e+4}\\
&(4.2587e+4)&(6.2467e+3)&(1.9218e+4)&(1.4402e+3)&(6.3779e+4)&(5.1452e+3)&(1.1582e+4)&(4.2843e+4)&\hl{(9.288e+3)}\\[1mm]

\multirow{2}{*}{$f_7$}&7.8310e+9 $-$&4.9258e+13 $-$&2.3870e+12 $-$&1.1084e+10 $-$&2.8185e+11 $-$&2.2211e+13 $-$&8.9123e+8 $\approx$&1.3523e+10 $-$&\hl{8.4166e+8}\\
&(2.9908e+9)&(2.3576e+13)&(7.2173e+11)&(9.4838e+9)&(1.4497e+11)&(8.4947e+12)&(1.9320e+8)&(1.6130e+9)&\hl{(7.997e+7)}\\[1mm]

\multirow{2}{*}{$f_8$}&2.9461e+16 $-$&1.6998e+17 $-$&\hl{1.2728e+13 $+$}&4.7204e+16 $-$&6.1725e+15 $\approx$&3.2323e+16 $-$&3.7727e+13 $+$&5.4599e+15 $\approx$&4.4372e+15\\
&(1.4136e+16)&(9.1594e+16)&\hl{(1.9676e+12)}&(6.0409e+15)&(3.8102e+15)&(3.0414e+16)&(3.0187e+13)&(2.5604e+15)&(1.770e+15)\\[1mm]

\multirow{2}{*}{$f_9$}&7.5728e+8 $-$&2.5298e+9 $-$&1.2725e+9 $-$&8.4502e+8 $-$&1.6577e+9 $-$&2.0244e+9 $-$&5.8016e+8 $-$&7.7668e+8 $-$&\hl{4.4601e+8}\\
&(1.2373e+8)&(2.0469e+8)&(6.3827e+7)&(4.5495e+7)&(1.3394e+8)&(2.1854e+8)&(1.1320e+8)&(4.1594e+7)&\hl{(8.503e+7)}\\[1mm]

\multirow{2}{*}{$f_{10}$}&4.4221e+7 $-$&8.7041e+7 $-$&2.1046e+6 $-$&9.6156e+7 $-$&1.7440e+7 $-$&7.6458e+7 $-$&1.2506e+6 $-$&7.8125e+5 $\approx$&\hl{4.8728e+5}\\
&(1.3077e+7)&(2.6524e+6)&(6.4391e+5)&(4.6215e+5)&(3.3440e+6)&(5.2418e+6)&(1.4290e+4)&(2.8894e+5)&\hl{(4.870e+5)}\\[1mm]

\multirow{2}{*}{$f_{11}$}&7.4972e+11 $-$&3.0398e+15 $-$&1.1061e+14 $-$&4.0919e+11 $-$&1.1334e+13 $-$&8.9529e+14 $-$&\hl{8.7764e+9 $+$}&4.4608e+11 $-$&1.6943e+11\\
&(4.5447e+11)&(1.4680e+15)&(9.8496e+13)&(1.0871e+11)&(5.1094e+12)&(3.6702e+14)&\hl{(4.5005e+9)}&(1.6708e+11)&(1.144e+11)\\[1mm]

\multirow{2}{*}{$f_{12}$}&2.0599e+10 $-$&1.9723e+12 $-$&1.5925e+12 $-$&1.3548e+10 $-$&2.6665e+12 $-$&1.6723e+12 $-$&3.3481e+10 $-$&6.4235e+11 $-$&\hl{7.6900e+8}\\
&(3.9186e+9)&(8.1522e+10)&(3.3820e+10)&(1.3457e+10)&(1.0994e+11)&(2.5065e+10)&(5.8270e+9)&(1.9268e+10)&\hl{(2.144e+7)}\\[1mm]

\multirow{2}{*}{$f_{13}$}&4.0277e+10 $-$&1.9190e+15 $-$&5.4725e+13 $-$&8.3330e+10 $-$&3.0193e+12 $-$&1.0082e+15 $-$&\hl{1.0448e+10 $+$}&6.6059e+10 $-$&1.7051e+10\\
&(1.4024e+10)&(8.0609e+14)&(1.7016e+13)&(1.4211e+10)&(1.6034e+12)&(6.2874e+14)&\hl{(2.3651e+9)}&(9.5957e+9)&(4.590e+9)\\[1mm]

\multirow{2}{*}{$f_{14}$}&1.0031e+12 $-$&4.5075e+15 $-$&1.2185e+14 $-$&9.6175e+11 $-$&1.2889e+13 $-$&1.1075e+15 $-$&\hl{7.7424e+10 $+$}&7.2323e+11 $-$&2.3734e+11\\
&(3.8285e+11)&(3.3588e+15)&(7.0856e+13)&(4.0377e+11)&(4.1028e+12)&(4.3764e+14)&\hl{(1.3029e+10)}&(8.4437e+10)&(7.915e+10)\\[1mm]

\multirow{2}{*}{$f_{15}$}&5.7586e+7 $-$&1.8722e+15 $-$&3.4395e+14 $-$&9.9936e+8 $-$&8.8170e+13 $-$&8.7705e+14 $-$&3.0449e+7 $\approx$&6.1844e+12 $-$&\hl{2.7840e+7}\\
&(1.5169e+7)&(2.4943e+14)&(4.9325e+13)&(1.1950e+9)&(2.5655e+13)&(1.2736e+14)&(4.6006e+6)&(1.4191e+12)&\hl{(2.630e+6)}\\[1mm]
\hline
\multicolumn{1}{c}{$+/-/\approx$}&1/14/0&0/15/0&1/14/0&1/14/0&0/14/1&0/15/0&5/8/2&0/13/2&\\
\bottomrule
\end{tabular}
}
\begin{tablenotes}
\footnotesize
\item["\dag"] '$+$', '$-$' and '$\approx$' indicate that the result is significantly better, significantly worse, and statistically similar to that obtained by AutoV.
\end{tablenotes}
\end{table*}

While the 13 benchmark problems are already translated and scaled, they are further rotated by randomly generated orthogonal matrices and challenge the nine metaheuristics. As can be seen from the experimental results listed in Table~\ref{tab:exp2}, the superiority of the proposed AutoV becomes more significant, where AutoV outperforms the other metaheuristics on 11 out of 13 problems. As a consequence, the superiority of AutoV over some classical and state-of-the-art metaheuristics can be verified. Besides, it also implies that AutoV is more effective than the approaches based on the recommendation and combination of existing metaheuristics, whose performance can hardly go beyond the best existing metaheuristic on each problem.

\subsection{Comparison on Large-Scale Benchmark Problems}

Lastly, the proposed AutoV and the eight existing metaheuristics are compared on the 15 CEC'2013 large-scale benchmark problems \cite{li2013benchmark}. These benchmark problems contain approximately 1000 decision variables and a variety of landscape functions, transformations, and interactions between variables, posing stiff challenges to general metaheuristics. For all the compared algorithms, the population size is set to 100 and the number of function evaluations is set to 120000.

Table~\ref{tab:exp3} presents the means and standard deviations of the minimum objective values found by the compared metaheuristics, averaged over 30 runs. It can be observed from the statistical results that the proposed AutoV also exhibits better overall performance than the others on the large-scale benchmark problems, achieving the best results on 9 out of 15 problems. It is noteworthy that although SHADE and IMODE suggest many complex search strategies for the combination of multiple operators and adaptation of parameters, they are still underperformed by AutoV that only contains a simple operator designed automatically. Therefore, the proposed AutoV offers bright prospects to the design of metaheuristics, which can potentially replace the laborious manual design process.

\section{Conclusions}

To reduce the human expertise in designing metaheuristics, this paper has analyzed the importance of translation, scale, and rotation invariance to the robustness of operators, and deduced the generic form of translation, scale, rotation invariant operators. Based on the deduced generic form, this paper has proposed a principled approach to search for high-performance operators automatically. In contrast to the automated design approaches based on existing metaheuristics, the proposed approach does not rely on any existing techniques, and can obtain competitive performance over some state-of-the-art metaheuristics on complex and large-scale optimization problems.

The experimental results have demonstrated the effectiveness and potential of the automated design of variation operators, and further research on this topic is highly desirable. Firstly, it is reasonable to search for high-performance operators based on more complex functions (e.g., include more parents and consider the update of velocity), where more effective operators are expected to be found. Secondly, since the proposed approach searches for operators according to their performance on a benchmark problem, it is reasonable to adopt more representative and practical problems to find high-performance operators for general problems or specific types of problems. Thirdly, it is interesting to develop novel approaches for automatically designing selection strategies for metaheuristics.

\bibliographystyle{IEEEtran}
\bibliography{AutoV}

\end{document}